\documentclass{article}

\usepackage{arxiv}

\usepackage[utf8]{inputenc} % allow utf-8 input
\usepackage[T1]{fontenc}    % use 8-bit T1 fonts
\usepackage{hyperref}       % hyperlinks
\usepackage{url}            % simple URL typesetting
\usepackage{booktabs}       % professional-quality tables
\usepackage{amsfonts}       % blackboard math symbols
\usepackage{nicefrac}       % compact symbols for 1/2, etc.
\usepackage{microtype}      % microtypography
\usepackage{graphicx}
\usepackage{doi}

\usepackage{amsmath,amssymb,booktabs,tabularx,algorithm,algpseudocode,graphicx}
\usepackage{tikz}
\usetikzlibrary{arrows.meta,positioning,shapes.geometric,fit}
\usepackage{array,enumitem}
\usepackage[numbers,sort&compress]{natbib}

\title{\textbf{AgentBetta: Verification-Driven Adaptive Configuration of an AI Nano-Agent through Selective Expansion and Verified Contraction}}

\author{%
  Md. Ashraful Babu$^{1,2}$ \\
  \normalfont $^1$Department of Physical Sciences\\
  \normalfont $^2$Explainable Artificial Intelligence and Optimization Research (XAiOR) \\
	Independent University, Bangladesh\\
	Dhaka, 1245, Bangladesh. \\
	\texttt{ashraful388@gmail.com; ashraful388@iub.edu.bd} \\
}

\renewcommand{\headeright}{}
\renewcommand{\undertitle}{}
\renewcommand{\shorttitle}{\textit{AgentBetta}: an adaptive AI Nano-Agent framework}

\hypersetup{
pdftitle={AgentBetta: Verification-Driven Adaptive Configuration of an AI Nano-Agent},
pdfsubject={Adaptive AI Agents and Large Language Model Systems},
pdfauthor={Md. Ashraful Babu},
pdfkeywords={adaptive AI agents, Nano-Agent, capability configuration, verification-driven adaptation},
}

\begin{document}
\maketitle

\begin{abstract}
Large language model agents are typically deployed with predefined configurations, although the required model capability, context, tools, permissions, memory, and computational resources can vary substantially across tasks. This study develops and evaluates AgentBetta, an adaptive AI Nano-Agent framework that represents these factors as an executable configuration and updates them through verification-driven diagnosis, selective expansion, and verification-based counterfactual contraction. The evaluation distinguishes controlled mechanism validation from external agent comparisons. On the AB-ConfigBench benchmark, AgentBetta achieved 91.38\% verified success while reducing median context allocation from 64,000 to 8,000 context characters and median tool exposure from five tools to zero compared with the fully provisioned configuration. The configuration-deficiency diagnosis achieved a macro-F1 score of 0.819 with precision of 1.000 across the evaluated dimensions, and selective expansion avoided unnecessary changes to unrelated configuration dimensions. Post-success contraction preserved verification outcomes in 56.41\% of evaluated one-dimension contraction probes, indicating that some successful configurations contained removable capability under the tested conditions. External evaluations indicate that adaptive configuration can improve the balance between verified task completion and capability exposure; however, the results vary across benchmarks and agent families. In particular, the cross-family replication did not reproduce the primary-backbone accuracy ordering, and specialized systems remained advantageous for certain task domains. These results support interpreting AgentBetta as a configuration-adaptation mechanism that regulates capability allocation and inference expenditure rather than as a universal replacement for specialized agent architectures.
\end{abstract}

% keywords can be removed
\keywords{adaptive AI agents \and Nano-Agent \and capability configuration \and verification-driven adaptation \and selective expansion \and counterfactual contraction\and resource-aware agent systems \and large language models.}
%%%%%%%%%%%%%%%%%%%%%%%%%%%%%%%%%%%%%%%%%

\section{Introduction}
\label{sec:introduction}

Large language models (LLMs) become operational agents when model inference is coupled with external actions, environmental observations, memory, and control logic. Early work established several elements that are now standard in agentic systems: interleaved reasoning and acting \cite{yao2023react}, learned use of external tools \cite{schick2023toolformer}, trial-to-trial improvement through linguistic feedback \cite{shinn2023reflexion}, and controller-based delegation to heterogeneous expert models \cite{shen2023hugginggpt}. These developments shifted the design problem from selecting a language model alone to constructing an executable system around the model. For a given task, performance can depend on the selected model, the information made visible to it, the available tools, the authority granted to those tools, the memory state, the resource budget, and the rules governing when execution should stop, retry, or escalate.

Most deployed agent systems nevertheless retain a substantial amount of configuration that is fixed before a task begins. A capable model may be exposed to a large context, a broad tool catalogue, permissive execution rights, persistent memory, and generous token or turn budgets even when the task requires only a small subset of these facilities. The converse problem also occurs: a deliberately economical configuration may omit information or capability required by a more demanding task. The resulting tension is not only between accuracy and monetary cost. Excess context can dilute relevant evidence, large toolsets can increase selection ambiguity, broad authorization enlarges the action surface, and unnecessary iterations increase both latency and the probability of execution error. Conversely, aggressive reduction can induce repeated retrieval, compensatory tool calls, or failure to complete the task. The central design question is therefore not whether an agent should always be ``small'' or always be ``powerful'', but how its executable configuration should be matched to the requirements of the current task and revised when those requirements become clearer during execution.

To investigate this problem, this study develops and evaluates \emph{AgentBetta}, an adaptive AI Nano-Agent based on verification-driven configuration control. The term \emph{Nano-Agent} is used here as an operational descriptor rather than as a claim of a new agent category: AgentBetta is intended to begin with a task-appropriate, bounded configuration and to acquire additional capability only when the execution evidence justifies it. A demanding task may consequently use a high-capability model, a larger context, additional tools, or longer execution limits; the ``nano'' property refers to avoiding unjustified allocation rather than to enforcing a permanently small model or a fixed resource ceiling. The executable configuration is represented as
\begin{equation}
X=
\left(M,C,T,P,\mathrm{Mem},R,\tau,I\right),
\label{eq:intro_configuration}
\end{equation}
where $M$ denotes model capability, $C$ the model-visible context, $T$ the exposed tools, $P$ the permission or authority set, $\mathrm{Mem}$ the available memory, $R$ quantitative computational or monetary resources, $\tau$ time/turn/retry bounds, and $I$ other interaction limits. AgentBetta treats $X$ as a task-conditioned control object rather than as a static deployment choice.

\subsection{Related literature and positioning}
\label{subsec:intro_literature}

The literature has already established that agent structure and resource allocation materially affect both task performance and execution cost. Automated workflow design methods make the control structure itself an optimization variable. AFlow searches over code-represented workflows using Monte Carlo Tree Search and execution feedback, showing that automatically discovered workflows can improve task performance while reducing reliance on manual design \cite{zhang2025aflow}. Complementary empirical work on efficient agents examines the effect of backbone choice, planning, tool use, memory, and test-time scaling on the cost--performance trade-off, demonstrating that additional agent components do not contribute uniformly across tasks \cite{wang2025efficientagents}. AgentSlimming addresses overprovisioned multi-agent workflows directly by pruning agent nodes and replacing expensive node models with lower-cost alternatives under a validation rule designed to avoid unacceptable performance loss \cite{chen2026agentslimming}. These studies establish that agentic systems can contain substantial structural and computational redundancy; they also show that efficiency cannot be inferred from task success alone.

A complementary body of work addresses input-dependent agent configuration. ARC formulates query-wise agent configuration as a semi-Markov decision process and learns a hierarchical policy over workflow structure, tools, token budgets, and prompt fragments \cite{taparia2026arc}. AOrchestra represents a dynamically instantiated sub-agent as a tuple comprising instruction, context, tools, and model, enabling a central orchestrator to create task-specific executors during a long-horizon trajectory \cite{ruan2026aorchestra}. ToolOrchestra trains a lightweight orchestrator to choose among heterogeneous tools and stronger language models while optimizing outcome quality, efficiency, and user tool preferences \cite{su2026toolorchestra}. AgentFlow instead places learning inside the multi-turn execution loop: a planner, executor, verifier, and generator interact through evolving memory, while the planner is optimized from trajectory-level outcome feedback \cite{li2026agentflow}. Eureka extends task-conditioned architecture formation further by constructing dynamic obligation graphs and specialized macro-agents with local state, memory, operators, tools, verifiers, and topology for long-horizon scientific tasks \cite{wong2026eureka}. Collectively, these studies establish dynamic model selection, tool selection, context allocation, and task-conditioned architecture formation as existing research directions.

Recent work also addresses the amount of execution that a task warrants. E3 formalizes minimum-sufficient execution and uses an Estimate--Execute--Expand procedure: it predicts an initial operating point, attempts a minimum viable path, and widens execution scope when verification fails \cite{yin2026e3}. This result is especially relevant to AgentBetta because it establishes that a lean initial execution followed by verification-driven expansion is already a defined research direction. Accordingly, the present contribution does not lie in the general principle of ``starting small'' or in escalation after failure. The remaining methodological question is whether failure evidence can be attributed to particular dimensions of a heterogeneous agent configuration and whether only those dimensions can be expanded without silently increasing unrelated capability.

Individual configuration dimensions have also been studied in depth. ToolScope reduces ambiguity and prompt burden by merging overlapping tools and retrieving a task-relevant subset from large tool libraries \cite{liu2026toolscope}. TRACER learns per-tool context-retention ratios and explicitly prices downstream tool re-invocation, showing that context reduction can create hidden reacquisition costs even when the immediate compression appears beneficial \cite{lin2026tracer}. Resource governance is formalized by Agent Contracts, which defines multidimensional resource and temporal bounds together with success and termination criteria and conservation constraints for delegated budgets \cite{ye2026agentcontracts}. Authorization research addresses a different but related form of excess capability. PAuth derives task-scoped operation constraints from natural-language requests and checks whether concrete operations are implied by the task \cite{sharma2026pauth}, whereas CAPMAS maps natural-language requests to bounded privilege sets and propagates attenuable capability tokens through multi-agent delegation \cite{veski2026capmas}. These studies show that context, tools, resources, and permissions each admit specialized control mechanisms. They also indicate that efficiency and authority exposure should not be treated as interchangeable quantities: a low-cost configuration may still be overprivileged, and a tightly authorized configuration may still waste model or context resources.

Recent studies have also examined adaptation after execution evidence becomes available. CHILL-Harness formulates harness adaptation as a causal intervention problem and learns when alternative orchestration workflows have sufficient expected advantage to justify intervention \cite{fu2026chill}. AgentSlimming validates successive structural reductions against a performance threshold \cite{chen2026agentslimming}. ClawTrace records cost-attributed execution traces and uses them to derive preserve, prune, and repair rules; notably, its prune rules are extracted from successful trajectories rather than only from failures \cite{yuan2026clawtrace}. TRACER uses counterfactual consequence attribution at the level of retained tool outputs \cite{lin2026tracer}. Thus, counterfactual analysis, pruning after successful behavior, and cost-aware adaptation are themselves established ideas. A remaining methodological gap is whether these mechanisms can be applied to the \emph{whole executable configuration} within a bidirectional runtime process: failure-side diagnosis and selective expansion followed, when appropriate, by success-side contraction and independent re-verification.

Table~\ref{tab:intro_positioning} summarizes the closest methodological relationships. The comparison is intentionally stated in terms of each method's primary optimization object; it is not intended to imply that a listed system is incapable of auxiliary behaviors outside its principal scope.

\begin{table*}[t]
\centering
\caption{Positioning of AgentBetta relative to closely related adaptive-agent methods. The entries describe each method's primary optimization scope rather than every auxiliary capability of the corresponding implementation.}
\label{tab:intro_positioning}
\footnotesize
\begin{tabularx}{\textwidth}{@{}p{2.35cm}p{3.0cm}p{4.05cm}X@{}}
\toprule
Method & Primary adaptive object & Runtime feedback or adaptation & Principal distinction from the present study \\
\midrule
ARC \cite{taparia2026arc}
& Workflow, tools, budget, prompts
& Learns query-wise configurations from correctness and cost
& Does not make post-failure configuration-dimension diagnosis and verified post-success contraction the central runtime loop. \\

E3 \cite{yin2026e3}
& Execution scope
& Verification failure triggers progressive scope expansion
& Establishes lean execution and recovery, but primarily varies execution footprint rather than a general heterogeneous configuration. \\

AOrchestra \cite{ruan2026aorchestra}
& Instruction, context, tools, model
& Creates specialized sub-agents during execution and learns orchestration from trajectories
& Establishes runtime model/context/tool specialization; permissions, resource bounds, and configuration contraction are outside its principal abstraction. \\

AgentFlow \cite{li2026agentflow}
& Planning and tool-use policy
& Verifier signals and final outcomes support in-flow policy learning
& Verification-driven learning is central, whereas the full executable configuration is not the primary optimization variable. \\

CHILL-Harness \cite{fu2026chill}
& Harness workflow interventions
& Counterfactual effect estimates authorize advantageous workflow changes
& Establishes counterfactual harness control rather than explicit necessity testing over model, context, tools, permissions, memory, and resources. \\

AgentSlimming \cite{chen2026agentslimming}
& Multi-agent workflow nodes and node models
& Candidate pruning/replacement is re-evaluated against an acceptance threshold
& Establishes validated structural contraction, mainly at workflow/task level rather than as per-instance bidirectional configuration adaptation. \\

Agent Contracts / PAuth / CAPMAS \cite{ye2026agentcontracts,sharma2026pauth,veski2026capmas}
& Resources and authorization
& Contracts and authorization checks impose execution bounds
& Establish governance mechanisms that an adaptive controller must respect; they do not jointly optimize the broader executable configuration. \\

AgentBetta (proposed)
& Model, context, tools, permissions, memory, resources, and execution/interaction bounds
& Verification produces structured insufficiency evidence; controlled success-side trials update configuration evidence
& Integrates heterogeneous configuration control, dimension-selective expansion, verified contraction, policy boundaries, and history-based initialization within one runtime formulation. \\
\bottomrule
\end{tabularx}
\end{table*}

\subsection{Research gap}
\label{subsec:intro_gap}

The reviewed literature leaves a narrower gap than the general problem of adaptive agents. First, configuration optimization remains distributed across partially overlapping research threads. ARC and AOrchestra cover several important dimensions, whereas context-management, tool-filtering, resource-governance, and authorization methods typically optimize their respective dimensions separately. A task, however, can fail because of interactions among these dimensions. A weak model with complete evidence, a strong model with incomplete evidence, a correctly selected tool without sufficient authorization, and an adequate configuration constrained by an insufficient turn budget can all produce an unsuccessful run for different reasons. Treating these cases as the same generic failure obscures the control decision that should follow.

Second, verification is commonly used to determine whether another attempt is required, but a binary failure signal does not by itself identify \emph{which} part of the configuration is deficient. Scope expansion in E3 is a principled recovery mechanism \cite{yin2026e3}; AgentFlow uses verifier feedback to improve action planning \cite{li2026agentflow}; CHILL-Harness evaluates alternative workflow interventions \cite{fu2026chill}. What remains insufficiently characterized is a configuration-level diagnostic map
\begin{equation}
D(f)\subseteq
\left\{M,C,T,P,\mathrm{Mem},R,\tau,I\right\},
\label{eq:intro_diagnosis}
\end{equation}
where $D(f)$ denotes the subset of configuration dimensions implicated by the evidence associated with failure or indeterminate verification. Such a representation enables a direct experimental comparison between \emph{dimension-selective expansion} and broader escalation policies. The distinction is consequential: increasing model capability when the actual deficiency is authorization or missing evidence can increase cost without correcting the cause of failure.

Third, expansion and contraction are usually studied as separate optimization problems. E3 expands execution scope after failure \cite{yin2026e3}; AgentSlimming removes agent nodes or substitutes cheaper models from an existing workflow \cite{chen2026agentslimming}; ClawTrace derives efficiency-oriented prune rules from successful traces \cite{yuan2026clawtrace}; and CHILL-Harness evaluates counterfactual workflow interventions \cite{fu2026chill}. These approaches motivate, but do not by themselves establish, a unified runtime process in which the same configuration representation supports both directions of adaptation. A successful configuration is feasible, but success does not establish that every allocated capability was necessary. Conversely, the fact that a tool, context item, permission, or larger model was not visibly used in one trace is not sufficient evidence that it can be removed. Necessity must be tested under an explicit counterfactual protocol.

Fourth, cross-task adaptation often learns prompts, skills, routes, or workflows rather than an empirical relationship between task characteristics and configurations that have been independently verified under a declared protocol. This distinction becomes important when configuration dimensions have different cost and risk semantics. For example, reducing the context window and removing write permission are not commensurate interventions, although both reduce exposure. A reusable configuration history should therefore retain the task features, exact configuration, adaptation sequence, verification outcome, resource measurements, and unsuccessful contraction trials. Such a history can support more informed initialization, but a previously verified configuration is not necessarily minimal: unless contraction evidence is available, successful historical configurations may remain overprovisioned. Failed contraction trials are therefore important because omitting them would systematically bias the empirical frontier toward configurations that appear cheaper only because contrary evidence was discarded.

Adaptation itself incurs computational and operational overhead. Failure diagnosis, additional verifier calls, re-execution, and counterfactual contraction can consume more resources than they save on a one-off task. Recent context work similarly shows that an apparently economical reduction can transfer cost to later reacquisition \cite{lin2026tracer}. A practical adaptive controller therefore requires an explicit accounting of adaptation overhead and, for optional learning experiments, a criterion for whether the expected future benefit justifies the cost of collecting additional evidence.

To address these gaps, this study develops AgentBetta as a verification-driven, bidirectional configuration controller. The initial configuration is inferred from task characteristics and, when sufficiently supported, prior verified run evidence. If verification fails or remains indeterminate, structured diagnostic evidence is mapped to one or more configuration dimensions and only admissible deficient dimensions are expanded when possible. After verified success, controlled environments may be used to construct reduced counterfactual configurations; these configurations are re-executed and independently verified rather than being declared sufficient from trace inspection alone. Evidence from successful and unsuccessful expansion and contraction trials is retained as an empirical verified configuration frontier that can be consulted during subsequent initialization. This history mechanism is treated as a warm-start aid rather than as evidence that the selected starting configuration is minimal. Hard policy constraints remain outside the adaptive controller: prediction that additional authority would be useful can generate a request, but it cannot authorize that authority autonomously.

The key contributions of this study are as follows:

\begin{itemize}

\item \textbf{Development and implementation of a verification-driven adaptive AI Nano-Agent.}
This study develops and implements AgentBetta, an adaptive AI Nano-Agent that represents model capability, context exposure, tools, permissions, memory, resources, execution bounds, and interaction limits jointly as the executable configuration $X=(M,C,T,P,\mathrm{Mem},R,\tau,I)$. The configuration can be selectively revised in response to runtime verification evidence while remaining subject to explicit policy constraints.

\item \textbf{Verification-driven capability adaptation.}
AgentBetta introduces a task-conditioned adaptation loop that combines configuration diagnosis, selective expansion, execution verification, and evidence-based capability adjustment. The adaptation process is constrained by predefined policies rather than unrestricted model decisions.

\item \textbf{Verification-guided contraction evaluation.}
Following verified task completion, AgentBetta constructs controlled single-dimension contraction candidates and accepts a reduction only when re-execution under the reduced configuration remains verified according to the declared evaluation protocol. The procedure evaluates removable capability exposure under controlled conditions rather than performing unrestricted automatic minimization.

\item \textbf{Capability exposure and resource-aware evaluation.}
The framework separately measures task success, capability exposure, inference expenditure, and adaptation overhead, enabling analysis of the trade-off between verified completion and unnecessary capability activation.

\item \textbf{Auditable adaptive agent execution.}
AgentBetta records configuration states, adaptation decisions, verification outcomes, and execution traces to support reproducible analysis of adaptive agent behavior.

\end{itemize}

The contribution claims concern the proposed formulation and implemented control mechanisms and do not imply global optimality or universal superiority over existing agents.

% AgentBetta manuscript methodology section only.
% Required packages in the parent manuscript:
% \usepackage{amsmath,amssymb,booktabs,tabularx,algorithm,algpseudocode,graphicx}
% The two methodology figures are supplied as external PNG files.

\section{Methodology}
\label{sec:methodology}

\subsection{Methodological scope and design objective}
\label{subsec:scope}

AgentBetta is formulated as an adaptive AI Nano-Agent whose executable configuration is treated as a task-dependent control variable rather than as a fixed deployment choice. The term \emph{Nano-Agent} denotes an operational principle: the runtime should allocate no more model capability, context, tools, authority, memory, or computational resources than are justified by the current task state and the evidence accumulated during execution. It does not imply that the underlying language model must be small, nor does it impose a permanently minimal configuration. A demanding task may therefore require a high-capability model, a larger evidence window, additional tools, or longer execution limits, provided that these additions are justified by verification evidence and remain within the applicable safety policy.

The method is designed around a bidirectional control process. Before execution, AgentBetta estimates a suitable initial configuration. During execution, task-specific verification is used to determine whether the current configuration is sufficient. When execution is unsuccessful or remains unverifiable, a diagnosis stage identifies the configuration dimensions that are plausibly responsible for the insufficiency, and only those dimensions are expanded where possible. After verified success, controlled counterfactual contraction may be performed in benchmark or learning settings to determine whether a smaller configuration preserves task success. Evidence from both expansion and contraction trials is retained for optional history-informed initialization of subsequent, related tasks. This reuse is treated as a warm-start mechanism whose effect on adaptation effort and starting exposure must be established empirically rather than assumed.

This formulation distinguishes the proposed control problem from the underlying agent loop itself. Interleaved reasoning, action, and observation are established components of language-agent systems \cite{yao2023react}, while recent work has separately examined query-conditioned agent configuration \cite{taparia2026arc}, dynamic sub-agent construction \cite{ruan2026aorchestra}, verification-driven scope expansion \cite{yin2026e3}, resource-bounded execution \cite{ye2026agentcontracts}, and workflow contraction \cite{chen2026agentslimming}. AgentBetta uses these developments as methodological context; the present method focuses on the coordinated adaptation of a heterogeneous configuration under explicit verification and policy constraints.

\begin{figure*}[t]
\centering
\includegraphics[width=0.98\textwidth]{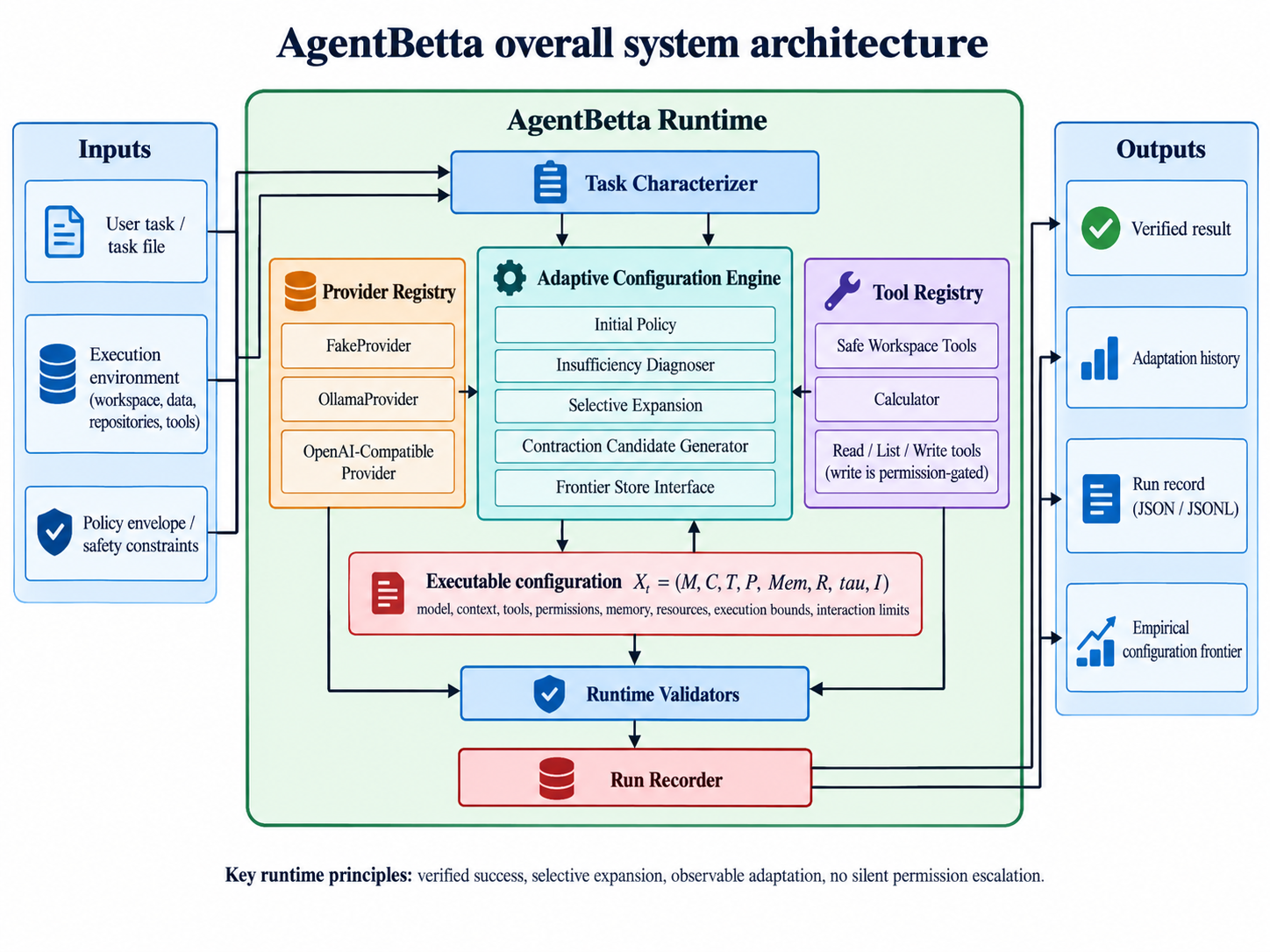}
\caption{Overall AgentBetta system architecture. The adaptive configuration engine maintains the executable configuration while provider and tool registries expose only policy-admissible capabilities. Runtime validators determine whether execution is acceptable, and run records preserve verification outcomes, adaptation history, and empirical configuration evidence.}
\label{fig:agentbetta_architecture}
\end{figure*}

Figure~\ref{fig:agentbetta_architecture} shows the implementation-level organization used to realize this formulation. Task characterization and policy constraints feed an adaptive configuration engine that coordinates provider selection, exposed tools, runtime verification, adaptation events, and run recording. The architecture deliberately separates provider availability, tool availability, configuration state, and verification so that a change in one dimension does not silently widen the others.

\subsection{Task model and executable configuration}
\label{subsec:configuration}

A task is represented as
\begin{equation}
q = \left(u,\,\mathcal{E},\,\mathcal{V},\,\Pi\right),
\label{eq:task}
\end{equation}
where $u$ is the user instruction, $\mathcal{E}$ denotes the accessible execution environment, $\mathcal{V}$ is the task-specific verification procedure, and $\Pi$ denotes immutable user and system policies that bound admissible actions. The environment may contain local files, structured data, software repositories, deterministic computation tools, or permitted external services. The verification procedure is defined independently of the adaptive controller wherever an objective evaluator is available.

At adaptation step $t$, the executable AgentBetta configuration is
\begin{equation}
X_t = \left(M_t,C_t,T_t,P_t,\mathrm{Mem}_t,R_t,\tau_t,I_t\right),
\label{eq:configuration}
\end{equation}
where each element corresponds to a measurable configuration dimension. Table~\ref{tab:configuration_dimensions} summarizes the operational interpretation of these dimensions. The representation is versioned so that every run can be reconstructed from its recorded task specification, configuration, provider state, and adaptation history.

\begin{table}[t]
\centering
\caption{Operational dimensions of the AgentBetta configuration.}
\label{tab:configuration_dimensions}
\small
\begin{tabularx}{\linewidth}{@{}p{1.35cm}p{2.6cm}X@{}}
\toprule
Symbol & Dimension & Operational interpretation \\
\midrule
$M$ & Model capability & Provider/model identity, capability tier, reasoning or modality support, and inference parameters that are under policy control. \\
$C$ & Context & Model-visible task information, retrieved evidence, history, and admissible context budget. \\
$T$ & Tools & Explicit set of callable tools or functions exposed to the agent for the current task. \\
$P$ & Permissions & Authorized operations and resource scopes; permission expansion can request additional authority but cannot grant it autonomously. \\
$\mathrm{Mem}$ & Memory & Working or persistent memory available to the task, together with retention and retrieval policy. \\
$R$ & Resources & Token, compute, monetary, or other quantitative resource budgets not represented elsewhere. \\
$\tau$ & Execution bounds & Time, turn, retry, and other stopping limits. \\
$I$ & Interaction limits & Bounds on communication, external interaction, or related execution controls when these are applicable. \\
\bottomrule
\end{tabularx}
\end{table}

For a task $q$, configuration selection is expressed as a constrained optimization problem,
\begin{equation}
X^{\star}(q)=\arg\min_{X\in\mathcal{F}(\Pi)} J(X;q),
\label{eq:objective}
\end{equation}
subject to
\begin{align}
\Pr\!\left(v=\textsc{pass}\mid X,q\right) &\geq \theta, \\
\mathrm{Risk}(X,q) &\leq \rho, \\
\mathrm{Cost}(X,q) &\leq B,
\label{eq:constraints}
\end{align}
where $\mathcal{F}(\Pi)$ is the set of configurations permitted by the immutable policy envelope $\Pi$, $\theta$ is the required success level, $\rho$ is a prescribed risk bound, and $B$ is the available resource budget. The objective $J$ is evaluated from measurable quantities. A general form is
\begin{equation}
J(X;q)=
\lambda_c\widetilde{C}_{\mathrm{mon}}
+\lambda_l\widetilde{L}
+\lambda_x\widetilde{E}_{C}
+\lambda_t\widetilde{E}_{T}
+\lambda_p\widetilde{E}_{P}
+\lambda_r\widetilde{R}
+\lambda_a\widetilde{A},
\label{eq:utility}
\end{equation}
where the normalized terms denote monetary cost, latency, context exposure, tool exposure, permission exposure, resource consumption, and adaptation overhead, respectively. The weights $\lambda_{\cdot}$ and normalization rules are fixed before evaluation on the held-out test set. Equation~\eqref{eq:objective} is used as a decision framework rather than as an assumption that a global optimum can be recovered from a finite number of executions.

\subsection{Task characterization and initial configuration}
\label{subsec:initialization}

AgentBetta first maps the task to a feature representation $\phi(q)$. Only information available before substantive execution is used for this step; oracle labels or hidden test outcomes are excluded. The feature representation can include task modality, input size, dependency cues, expected side-effect class, explicitly requested operations, available tool requirements, privacy or local-only constraints, and coarse indicators of structural or reasoning complexity. Features derived from environment inspection are restricted to low-cost probes permitted by $\Pi$.

The initial configuration is generated by
\begin{equation}
X_0 = \pi_0\!\left(\phi(q),\mathcal{H}\right),
\label{eq:initial_policy}
\end{equation}
where $\mathcal{H}$ is the accumulated configuration history. The reference implementation uses a deterministic initialization policy when no admissible historical evidence is available. When history-informed initialization is enabled, completed verified runs are indexed by a predeclared task-feature bucket, and configurations from compatible training/development evidence are ranked by a recorded resource score. The lowest-scoring supported verified configuration is used as the warm start; current task permissions remain authoritative, and write-gated tools are removed when the corresponding authority is absent. If no compatible verified record is available, initialization falls back deterministically. Because a stored successful configuration need not be minimal, history reuse is interpreted as a warm start rather than as proof that $X_0$ is the least-exposure feasible configuration. The basis for the selected $X_0$ is recorded in the run metadata.

The initial policy is deliberately conservative with respect to authority. A predicted need for a privileged operation is represented as a permission requirement or request; it is not interpreted as authorization. This separation is consistent with task-scoped authorization principles in recent agent-security work \cite{sharma2026pauth} and prevents the adaptive controller from turning capability prediction into an implicit privilege grant.

\subsection{Execution state and verification}
\label{subsec:verification}

Execution proceeds as a sequence of model decisions, admissible tool calls, environment observations, and verification checkpoints. The runtime state at step $t$ is written as
\begin{equation}
s_t=\left(q,X_t,H_t,b_t\right),
\end{equation}
where $H_t$ is the accumulated execution history and $b_t$ contains remaining resource and time budgets. The action policy operates only over tools, context, and permissions exposed by $X_t$; capabilities that are absent from the configuration are not silently made available during execution.

The verifier returns
\begin{equation}
\mathcal{V}(s_t,y_t)=\left(v_t,z_t\right),
\label{eq:verifier}
\end{equation}
where $y_t$ is the current task output, $v_t\in\{\textsc{pass},\textsc{fail},\textsc{indeterminate}\}$ is the verification state, and $z_t$ is structured diagnostic evidence. Verification is ordered by evidential strength. Deterministic environment checks, executable tests, exact-match conditions, schema validation, and reproducible numerical checks are used when available. When such evaluators are unavailable, a fixed external evaluator may be used under a predeclared protocol. In that case, the evaluator is separated from the acting policy, is blinded to the identity of the adaptation condition where feasible, and its prompt, model version, and sampling parameters are recorded. This design reduces the risk that the same model both produces and uncritically validates its own output.

A failed verifier does not automatically imply inadequate model capability. For example, the cause may be unavailable evidence, a missing tool, an authorization boundary, a resource limit, or an execution error. AgentBetta therefore separates verification from insufficiency diagnosis.

\subsection{Configuration-insufficiency diagnosis}
\label{subsec:diagnosis}

When $v_t\neq\textsc{pass}$, the diagnosis function
\begin{equation}
d_t = g\!\left(\phi(q),X_t,H_t,z_t\right)
\label{eq:diagnosis}
\end{equation}
produces a ranked set of candidate deficiency labels with associated evidence. The diagnosis space is
\begin{equation}
\mathcal{D}=\{d_M,d_C,d_T,d_P,d_{\mathrm{Mem}},d_R,d_{\tau},d_I,d_{\varnothing}\},
\end{equation}
where $d_{\varnothing}$ denotes a non-configuration failure or an unresolved diagnosis. Operational signals include provider errors, context retrieval failures, explicit missing-tool conditions, permission denials, memory discontinuity, resource exhaustion, timeout or retry exhaustion, and verifier-specific evidence indicating that relevant information or capability was unavailable.

The production diagnosis uses only evidence observable at runtime. Ground-truth deficiency labels, when constructed for controlled experiments, are used solely to evaluate the accuracy of $g$ and are not supplied to the acting agent. The diagnosis output is stored as an \emph{AdaptationEvent} containing the pre-adaptation configuration, evidence, selected deficiency class or classes, confidence, proposed modification, and subsequent verification outcome. If no diagnosis exceeds the predeclared confidence requirement, the event is labelled unresolved; AgentBetta may perform a bounded diagnostic probe, apply a predeclared fallback policy, or terminate without claiming that a specific dimension was deficient.

\subsection{Dimension-selective configuration expansion}
\label{subsec:expansion}

Given a diagnosed set $D_t\subseteq\mathcal{D}\setminus\{d_{\varnothing}\}$, the expansion operator produces
\begin{equation}
X_{t+1}=\Gamma^{+}_{D_t}(X_t),
\label{eq:expansion}
\end{equation}
subject to the requirement that dimensions not included in $D_t$ remain unchanged unless a documented dependency makes that impossible. For a single diagnosed dimension $k$, the intended update therefore satisfies
\begin{equation}
X_{t+1}^{(j)}=X_t^{(j)} \quad \forall j\neq k,
\end{equation}
while $X_{t+1}^{(k)}$ is replaced by the next admissible capability level under the predeclared ordering for that dimension. Examples include increasing model capability without adding tools, expanding evidence context without changing permissions, adding one task-relevant tool without broad shell access, increasing an execution limit without changing the model, or enabling additional memory only when state continuity is implicated.

Expansion after unsuccessful verification has clear precedent in complexity-aware execution strategies such as E3 \cite{yin2026e3}. AgentBetta differs methodologically by treating expansion as an operation over a heterogeneous configuration and by recording which dimension was changed and why. The experiment therefore evaluates selective expansion against broad or wholesale escalation rather than assuming that selective adaptation is beneficial.

Permission adaptation is handled differently from ordinary resource expansion. If $d_P$ is diagnosed, $\Gamma^{+}_{D_t}$ may generate the narrowest permission request consistent with the failed operation, but the configuration does not acquire the requested authority until an external policy mechanism authorizes it. Hard-denied operations are not escalatable. Similarly, local-only or privacy restrictions cannot be relaxed by the adaptive policy. Resource bounds are treated as explicit constraints, consistent with the broader principle of auditable resource-bounded agent execution \cite{ye2026agentcontracts}.

\subsection{Post-success counterfactual configuration contraction}
\label{subsec:contraction}

A verified successful configuration establishes feasibility but does not establish that every allocated capability was necessary. AgentBetta therefore includes a counterfactual contraction procedure for controlled benchmark and learning settings. Let $X_s$ denote a configuration for which $v_s=\textsc{pass}$. For an admissible dimension $k$, a one-step contraction operator $\Gamma^{-}_k$ constructs
\begin{equation}
X_s^{-k}=\Gamma^{-}_k(X_s),
\end{equation}
where only dimension $k$ is reduced according to its predefined partial order. Candidate operations include selecting a lower-cost or lower-capability model, reducing active context, removing a tool, narrowing a permission set, reducing memory, lowering token or compute budgets, or reducing time/turn limits.

Each candidate reduction is evaluated by re-executing the task under the reduced configuration whenever such replay is safe and reproducible. For deterministic tasks, a single exact re-execution may be sufficient under the declared evaluator. For stochastic model executions, $n_c$ repeated trials are conducted under a fixed sampling protocol; paired random seeds are used when the provider exposes reproducible seeding. A contraction is retained only when the lower confidence bound on its empirical success probability satisfies the predeclared success threshold,
\begin{equation}
\mathrm{LCB}_{1-\alpha}\!\left(\widehat{p}_{k}\right)\geq\theta,
\label{eq:contraction_acceptance}
\end{equation}
and the reduced configuration satisfies all safety and budget constraints. The number of trials $n_c$, confidence level $1-\alpha$, and threshold $\theta$ are fixed before test-set evaluation.

Accepted contractions are applied sequentially, with the task re-verified after every accepted change. The process stops when no admissible one-step contraction passes the acceptance criterion, when the contraction budget is exhausted, or when further replay is unsafe. The resulting configuration is described as an \emph{empirically contracted feasible configuration}; it is not claimed to be a global minimum because the configuration space may be non-convex, categorical, stochastic, and only partially ordered. This terminology also distinguishes the procedure from workflow-level pruning methods such as AgentSlimming \cite{chen2026agentslimming}, which motivates the need for direct comparison rather than an assumption of methodological superiority.

Counterfactual contraction is never used to repeat irreversible or harmful real-world actions solely for experimental purposes. Such experiments are restricted to deterministic fixtures, sandboxes, simulations, replayable repositories, or other controlled environments. For live tasks, AgentBetta may record a contraction hypothesis, but it does not label a reduced configuration as sufficient unless the stated validation protocol has actually been satisfied.

\subsection{Learning the verified configuration frontier}
\label{subsec:frontier}

Every completed execution contributes a record
\begin{equation}
r_i=\left(\phi(q_i),X_i,\mathcal{A}_i,v_i,m_i\right),
\end{equation}
where $\mathcal{A}_i$ is the ordered adaptation history and $m_i$ contains measured cost, latency, context exposure, tool exposure, permission exposure, and resource use. Successful and unsuccessful contraction trials are both retained. Excluding failed contraction attempts would bias the estimated frontier toward configurations that appear artificially efficient.

For a task family or neighbourhood in feature space, the empirical feasible set contains configurations that satisfy the verification requirement under the declared protocol. A configuration $X_a$ dominates $X_b$ only when it is no worse on all prespecified optimization criteria and strictly better on at least one criterion while meeting the same success and safety constraints. The nondominated set defines an empirical \emph{verified configuration frontier}. This frontier is used to warm-start future tasks but is not treated as proof of universal optimality.

The implemented history layer is case based and deliberately conservative. A new task is mapped to the same predeclared feature representation used by the initialization policy, and only verified records from permitted data splits are eligible for reuse. Within a compatible feature bucket, the initializer selects the verified configuration with the lowest recorded resource score; when no eligible record exists, it returns to the deterministic initialization policy. Current authority is re-applied after retrieval, so historical permission state cannot widen the task's present authorization. This mechanism estimates an empirical warm start, not a minimal configuration. In particular, if the frontier contains only successful but relatively large configurations, history can reduce the number of adaptation steps without reducing initial capability exposure.

\subsection{Adaptation-cost control}
\label{subsec:adaptation_cost}

Diagnosis, re-execution, verification, and contraction consume resources. AgentBetta therefore distinguishes ordinary task-execution cost from the additional cost of collecting adaptation evidence. In benchmark mode, contraction follows a fixed experimental budget so that acceptance and rejection events can be compared consistently. For deployment, a natural decision criterion is to perform optional contraction only when its expected amortized benefit exceeds its estimated cost,
\begin{equation}
\widehat{N}_{\mathrm{reuse}}\,\widehat{\Delta J}_{\mathrm{future}}
>
\widehat{C}_{\mathrm{cf}}+\delta,
\label{eq:amortization}
\end{equation}
where $\widehat{N}_{\mathrm{reuse}}$ is the estimated number of future uses for the relevant task neighbourhood, $\widehat{\Delta J}_{\mathrm{future}}$ is the expected reduction in objective value per reuse, $\widehat{C}_{\mathrm{cf}}$ is the estimated cost of the counterfactual experiment, and $\delta$ is a non-negative safety margin. Equation~\eqref{eq:amortization} is a deployment decision criterion rather than an empirically established benefit in the reported benchmark campaign; the reported experiments account for adaptation activity separately and do not infer net efficiency from in-process fixture timing.

\subsection{Integrated control algorithm}
\label{subsec:algorithm}

\begin{figure*}[t]
\centering
\includegraphics[width=0.99\textwidth]{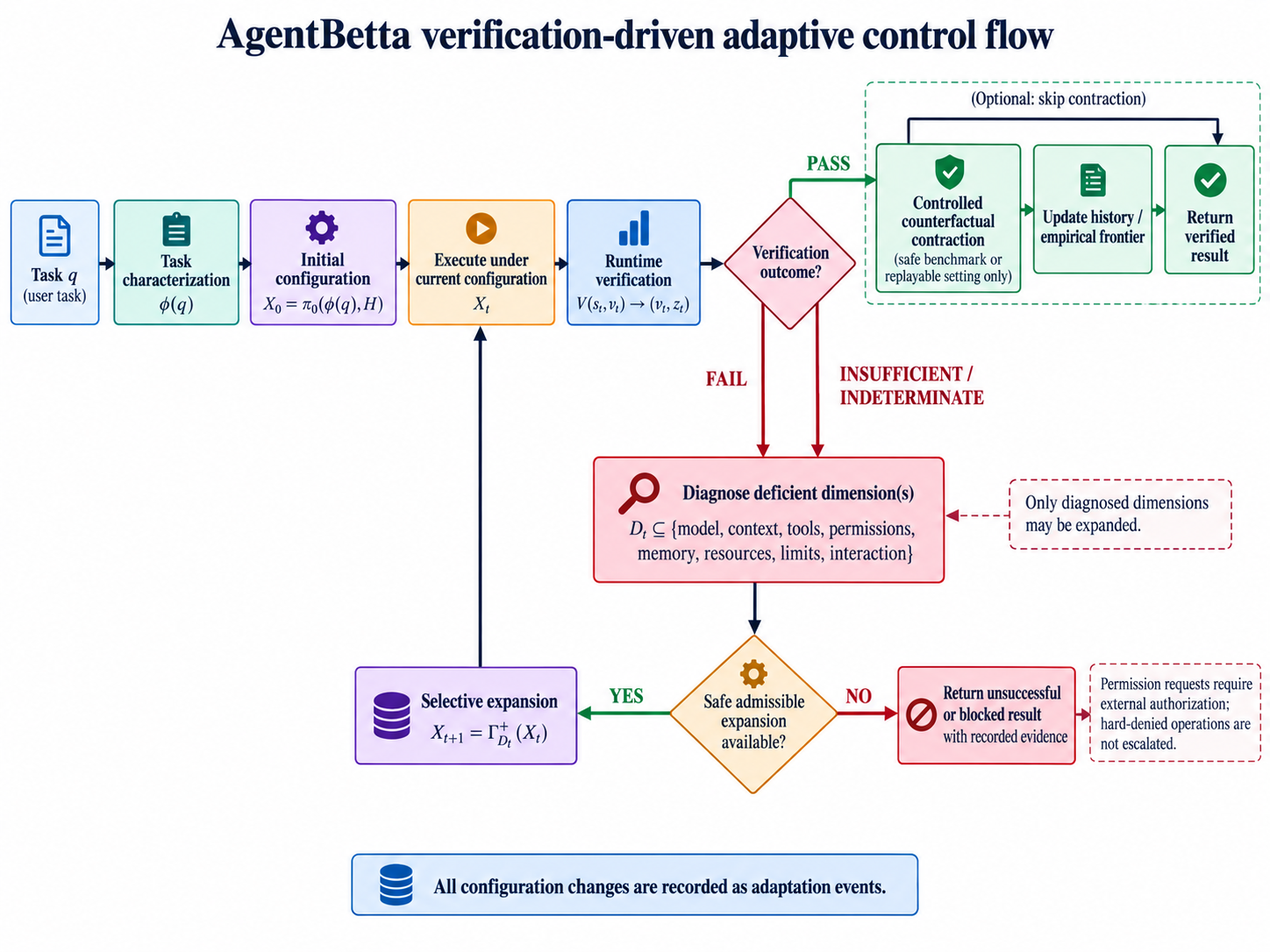}
\caption{Verification-driven adaptive control flow in AgentBetta. A task is characterized and assigned an initial configuration before execution and runtime verification. Unsuccessful or indeterminate verification triggers configuration-deficiency diagnosis; only policy-admissible diagnosed dimensions may be expanded. Verified configurations may undergo controlled counterfactual contraction in replayable settings, after which accepted evidence is recorded for subsequent history-informed initialization.}
\label{fig:agentbetta_loop}
\end{figure*}

Figure~\ref{fig:agentbetta_loop} summarizes the runtime logic, and Algorithm~\ref{alg:agentbetta} provides the corresponding control procedure. The acting model is never allowed to modify the hard policy envelope, verification definition, or experimental acceptance thresholds during a run.

\begin{algorithm}[t]
\caption{Verification-driven adaptive configuration in AgentBetta}
\label{alg:agentbetta}
\begin{algorithmic}[1]
\Require Task $q$, policy envelope $\Pi$, history $\mathcal{H}$, run budget $B$
\State $\phi \gets \textsc{Characterize}(q)$
\State $X \gets \pi_0(\phi,\mathcal{H})$
\State \textsc{ValidatePolicy}$(X,\Pi)$
\While{run budget and stopping conditions permit}
    \State $(y,H,m) \gets \textsc{Execute}(q,X)$
    \State $(v,z) \gets \mathcal{V}(q,X,y,H)$
    \State \textsc{RecordRunState}$(q,X,H,m,v,z)$
    \If{$v=\textsc{pass}$}
        \If{\textsc{ContractionEligible}$(q,X,m,\Pi)$}
            \State $(X_c,\mathcal{R}_c) \gets \textsc{CounterfactualContract}(q,X,\mathcal{V},\Pi)$
            \State \textsc{UpdateHistory}$(\mathcal{H},q,X_c,\mathcal{R}_c)$
        \Else
            \State \textsc{UpdateHistory}$(\mathcal{H},q,X,v,m)$
        \EndIf
        \State \Return verified result
    \EndIf
    \State $D \gets \textsc{Diagnose}(q,X,H,z)$
    \If{$D$ is unresolved or no safe admissible expansion exists}
        \State \Return unsuccessful/blocked result with recorded evidence
    \EndIf
    \State $X' \gets \textsc{SelectiveExpand}(X,D,\Pi)$
    \State \textsc{RecordAdaptation}$(X,D,X',z)$
    \State $X \gets X'$
\EndWhile
\State \Return limit-exhausted result with complete run record
\end{algorithmic}
\end{algorithm}

\subsection{Safety, authorization, and reproducibility constraints}
\label{subsec:safety_reproducibility}

The adaptive controller operates inside a non-adaptive safety envelope. User restrictions, hard-denied capabilities, workspace boundaries, credential protections, and local-only requirements are evaluated outside the language model and cannot be relaxed by model output. Permission state $P_t$ therefore represents granted authority, not the set of operations the model would prefer to perform. A proposed permission expansion creates a bounded request and remains ineffective unless the external authorization mechanism accepts it. This separation is important because task-specific authorization is itself an established security problem \cite{sharma2026pauth}; AgentBetta does not treat authorization enforcement as an emergent property of language-model reasoning.

For reproducibility, each run records the AgentBetta version and source revision; task identifier and task features; complete configuration sequence $X_0,\ldots,X_T$; provider and model identifiers; model parameters; tool and permission sets; input and output token counts when available; monetary cost where reliably measurable; wall-clock latency; tool invocations; verifier outputs; diagnosis evidence; adaptation events; stopping reason; and hashes of relevant artifacts when appropriate. Contraction trials are stored as first-class experimental observations, including failed trials. Random seeds are fixed when supported, and stochastic conditions are repeated according to the prespecified evaluation protocol.

The implementation exposes component controls for task-conditioned initialization, deficiency diagnosis, counterfactual contraction, and history-informed initialization; selective expansion is evaluated through controlled comparison with scope-only and wholesale expansion policies. These controls allow the empirical study to attribute observed changes to specific mechanisms rather than to a stronger underlying model, a larger tool set, or broader permissions. Adaptation overhead is recorded separately; the amortization criterion in Equation~\eqref{eq:amortization} is not treated as an experimentally validated gain unless it is explicitly activated and measured. The external evaluation further compares AgentBetta with fixed configurations and reasoning-, collaboration-, and specialization-oriented baselines under matched model, tool, verifier, and task conditions wherever technically feasible. Quantitative claims are made only from the frozen implementation and recorded experimental outputs.

\subsection{Experimental design and statistical analysis}
\label{subsec:experimental_design}

The empirical study was divided into a controlled mechanism-validation layer and an external model-facing layer. The primary experiments were executed on a Mac Studio (Apple M2 Max, 32~GB unified memory) running macOS 26.5.2. The external-evaluation environment used Python~3.13.9 and AgentBetta~0.2.0.dev0; repository-level SWE-bench reproduction used Python~3.11 for compatibility with the historical software environments represented by the selected instances. Run records preserve the source revision, benchmark and task identifiers, model and provider identifiers, budgets, verifier fingerprints, repetition index, resource measurements, and stopping status. The AgentBetta source code is publicly available at \url{https://github.com/ashraful388/AgentBetta}. For reproducibility, the software version associated with the experiments reported in this study is identified by its corresponding release tag and commit hash \cite{babu2026agentbetta}.

\paragraph{Controlled benchmark.}
AB-ConfigBench v1 contains 245 deterministic tasks. The benchmark includes 25 tasks targeting each of the eight configuration dimensions in Eq.~\eqref{eq:configuration}, together with 15 already-sufficient tasks and 10 tasks each in mixed-deficiency, impossible, and hard-denied categories. Each task stores a minimum admissible configuration and a ground-truth set of deficient dimensions; these labels are withheld from the acting policy and are used only for evaluation. A frozen positional split assigns 147 tasks to training, 40 to development, and 58 to the held-out test set. The headline held-out comparison in Table~\ref{tab:controlled_overall} uses the 58 test tasks. Broader mechanism analyses, including diagnosis and counterfactual-probe statistics, report their own denominators explicitly and are not interpreted as additional held-out test accuracy. Because the configuration-aware fixture provider is deterministic, this layer measures controller behavior rather than stochastic language-model performance.

\paragraph{External benchmarks and baselines.}
The principal external backbone was \texttt{qwen3.8-flash}, accessed through a fixed OpenAI-compatible model client with temperature~0. The same model client, native task tools, verifier, and benchmark-specific ceiling were supplied to all methods within a comparison. The core panel compared Fixed Lean, Fixed Full, matched ReAct, matched Reflexion, and AgentBetta. A separate 64-task BFCL subset compared AgentBetta with AutoGen~0.7.5 and CrewAI~1.15.21 using their published runtime abstractions. The coding pilot used SWE-agent~1.1.0. Cross-family replication used \texttt{llama3.1:8b} (Q4\_K\_M, on-device through Ollama) on a 16-task BFCL subset. These comparisons evaluate the implemented configurations under matched experimental conditions and should not be interpreted as exhaustive evaluations of the respective frameworks or their broader configuration spaces. The external data comprised BFCL v4 function-calling tasks, a 40-task text-only GAIA validation panel restricted to Levels~1--2, and eight screened SWE-bench Verified instances. The SWE-bench experiment used a declared local-checkout reproduction rather than the official Docker leaderboard harness; consequently, its results support only the within-study agent comparison. 
The BFCL core panel contained 191 shared tasks with three intended repetitions per method, and the AutoGen/CrewAI framework subset contained 64 tasks with three complete repetitions. GAIA used 40 tasks with three intended repetitions. The SWE-bench pilot used eight instances with two repetitions, and the cross-family BFCL panel used 16 tasks with three repetitions. Task selection and bootstrap sampling used the fixed seed 20260914. Benchmark-specific ceilings were shared within each comparison: BFCL allowed at most four model calls, eight tool calls, 24,000 context characters, and 600 output tokens per call; GAIA allowed 16 model calls, 24 tool calls, 64,000 context characters, and 1,200 output tokens per call; SWE-bench used a hard 40-model-call ceiling, 60 tool calls, 200,000 context characters, and 1,500 output tokens per call. The local cross-family BFCL panel used an eight-call ceiling, 12 tool calls, 32,000 context characters, and 512 output tokens per call.

\paragraph{Scoring and statistical analysis.}
BFCL function emissions were scored with the benchmark function-call checker; GAIA answers were evaluated under a fixed judge protocol; and SWE-bench success required the designated \texttt{FAIL\_TO\_PASS} tests to pass without regression of baseline-valid \texttt{PASS\_TO\_PASS} tests. For repeated external panels, the primary unit of analysis was the task. Scoreable repetitions were first averaged within each task for both success and continuous resource measurements, after which paired comparisons were formed on shared task identifiers. Pooled run-level success is reported only as a descriptive quantity. Provider or infrastructure failures that prevented scoring were excluded from the primary success denominator and are reported separately rather than imputed as task failures.

Binary paired comparisons used exact McNemar tests where applicable. Confidence intervals for paired task-level differences and continuous resource effects used a paired percentile bootstrap with 5,000 resamples and seed 20260914. Holm--Bonferroni correction was applied only to the predeclared confirmatory family comprising AgentBetta versus matched ReAct on BFCL, Reflexion on GAIA, the evaluated AutoGen and CrewAI configurations on the matched BFCL subset, and SWE-agent on the coding pilot. Non-significance is not interpreted as equivalence. Wall-clock latency is reported descriptively but is excluded from the principal efficiency claims because provider retry and network behavior were not uniformly bounded across methods.

\section{Results}
\label{sec:results}

\subsection{Evaluation scope and reporting convention}
\label{subsec:results_scope}

The evaluation follows the two-layer design and statistical protocol defined in Section~\ref{subsec:experimental_design}. The controlled AB-ConfigBench layer isolates configuration diagnosis, selective expansion, verification, contraction, and history reuse with a deterministic configuration-aware provider; it is therefore interpreted as mechanism validation rather than as a measure of general language-model capability. The external layer uses real model inference on BFCL \cite{patil2025bfcl}, a text-only GAIA panel \cite{mialon2024gaia}, and a screened SWE-bench Verified pilot \cite{jimenez2024swebench}, with matched reasoning-, reflection-, collaboration-, and specialization-oriented baselines. Primary repeated-panel comparisons use task-level repetition aggregates, while pooled run-level values are retained only as descriptive quantities. Unless otherwise specified, success throughout the evaluation refers to verifier-confirmed task completion under the declared benchmark-specific evaluation protocol.

\subsection{Controlled performance under task-conditioned configuration}
\label{subsec:controlled_overall}

Table~\ref{tab:controlled_overall} reports the held-out AB-ConfigBench test results. AgentBetta Full verified 91.38\% of the 58 tasks, compared with 93.10\% for Fixed Full and Wholesale Escalation. The small difference in verified success was accompanied by a substantial difference in provisioned capability. AgentBetta retained a median context allocation of 8,000 and a median tool exposure of zero, whereas Fixed Full and Wholesale Escalation used 64,000 context units and five exposed tools. Fixed Lean reached only 6.90\%, and removal of diagnosis reduced AgentBetta to 15.52\%. By contrast, removing contraction or history did not change the held-out success rate. The controlled results therefore indicate that the main success contribution arises from failure-side diagnosis and adaptation, while contraction and history primarily affect post-success exposure and search effort.

\begin{table}[t]
\centering
\caption{Held-out AB-ConfigBench performance under the deterministic configuration-aware provider. The experiment isolates controller behavior and is not a general LLM comparison.}
\label{tab:controlled_overall}
\small
\begin{tabularx}{\linewidth}{@{}Xrrrr@{}}
\toprule
Method & Verified success (\%) & Median context & Median tools & Median adaptations \\
\midrule
AgentBetta Full & 91.38 & 8,000 & 0 & 1.0 \\
Fixed Full & 93.10 & 64,000 & 5 & 0.0 \\
Wholesale Escalation & 93.10 & 64,000 & 5 & 1.0 \\
Fixed Lean & 6.90 & 8,000 & 0 & 0.0 \\
AgentBetta $-$Diagnosis & 15.52 & 8,000 & 0 & 0.0 \\
AgentBetta $-$Contraction & 91.38 & 8,000 & 0 & 1.0 \\
AgentBetta $-$History & 91.38 & 8,000 & 0 & 1.0 \\
\bottomrule
\end{tabularx}
\end{table}

Across the broader controlled campaign, the same pattern remained evident. Removing diagnosis reduced verified success from 89.67\% to 14.15\%, a decrease of 75.61 percentage points, with Holm-adjusted $p<10^{-45}$. Removing contraction or history changed success by only 0.09 percentage points and was not statistically distinguishable from the full method. Fixed Full and Wholesale Escalation did not show a statistically reliable success advantage over AgentBetta in the campaign-level paired analysis. These results separate the function of the components: diagnosis supports recovery before success, whereas contraction and history operate on the configuration search after or across successful runs.

\subsection{Configuration-deficiency diagnosis}
\label{subsec:diagnosis_results}

\begin{figure}[t]
\centering
\includegraphics[width=\linewidth]{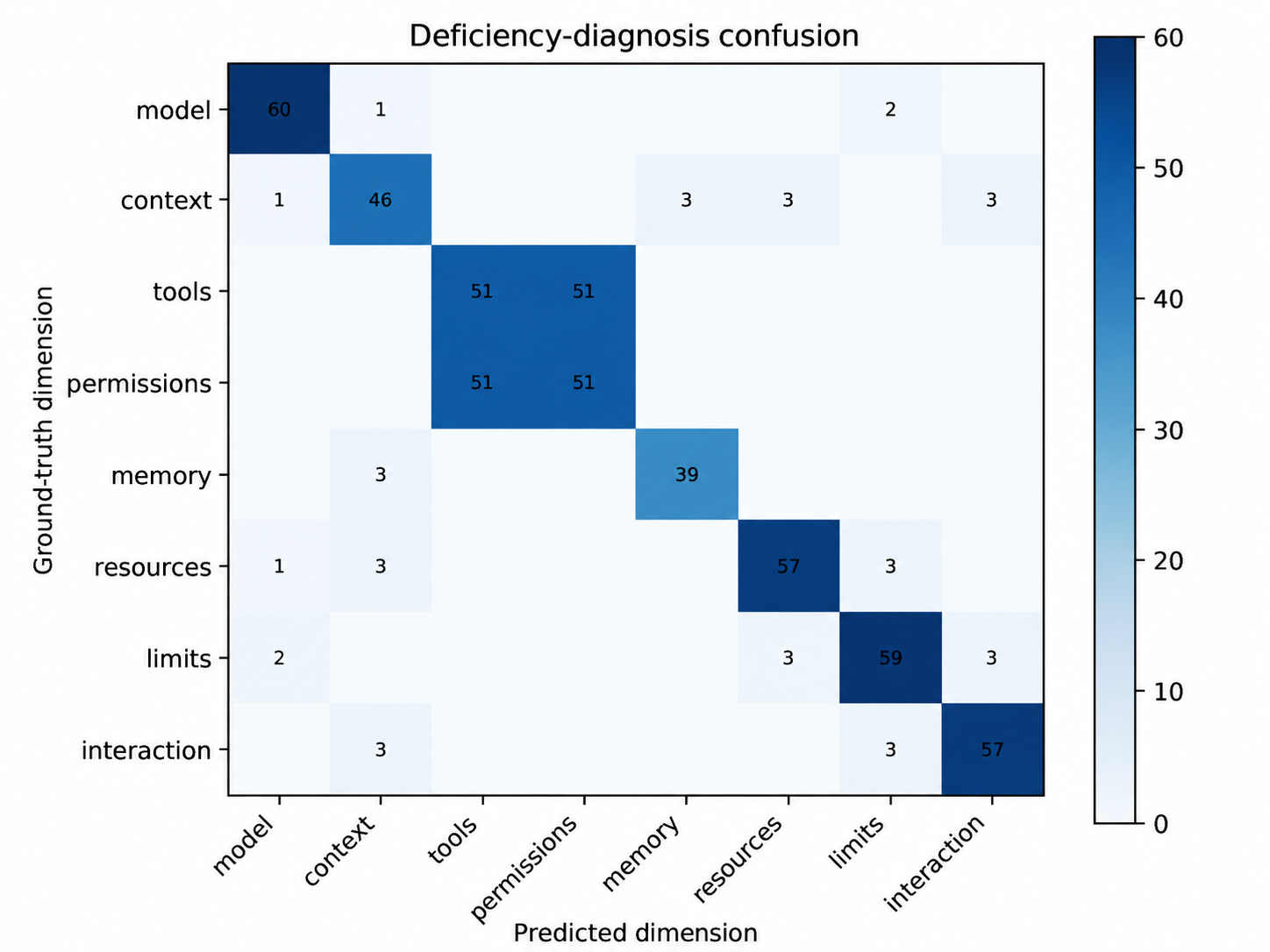}
\caption{Distribution of true and predicted configuration deficiencies in the controlled campaign. The diagonal concentration is consistent with the high precision reported in Table~\ref{tab:diagnosis}; residual error is dominated by missed rather than spurious deficiencies.}
\label{fig:diagnosis_heatmap}
\end{figure}

Configuration-deficiency diagnosis achieved macro-F1 0.819, micro-F1 0.800, exact-set accuracy 0.689, and Hamming loss 0.047. Precision was 1.000 for every configuration dimension (Table~\ref{tab:diagnosis}), so no evaluated diagnosis expanded a dimension absent from the ground-truth deficient set. The remaining errors were false negatives. Recall was highest for interaction limits (0.846), execution limits (0.840), and resources (0.818), and lowest for tools (0.354). The tool result was associated with deliberately retained tasks in which the required tool need was not inferable from the task information available to the controller. The diagnosis heatmap in Fig.~\ref{fig:diagnosis_heatmap} provides the corresponding distribution of predicted deficiencies.

\begin{table}[t]
\centering
\caption{Per-dimension configuration-deficiency diagnosis in the controlled campaign.}
\label{tab:diagnosis}
\small
\begin{tabular}{lrrrr}
\toprule
Dimension & Support & Precision & Recall & F1 \\
\midrule
Model & 218 & 1.000 & 0.716 & 0.834 \\
Context & 161 & 1.000 & 0.752 & 0.858 \\
Tools & 322 & 1.000 & 0.354 & 0.523 \\
Permissions & 187 & 1.000 & 0.610 & 0.758 \\
Memory & 142 & 1.000 & 0.739 & 0.850 \\
Resources & 154 & 1.000 & 0.818 & 0.900 \\
Execution limits & 156 & 1.000 & 0.840 & 0.913 \\
Interaction limits & 149 & 1.000 & 0.846 & 0.916 \\
\midrule
Macro average & -- & 1.000 & -- & 0.819 \\
\bottomrule
\end{tabular}
\end{table}

\subsection{Selective expansion versus wholesale escalation}
\label{subsec:selective_expansion_results}

\begin{figure}[t]
\centering
\includegraphics[width=\linewidth]{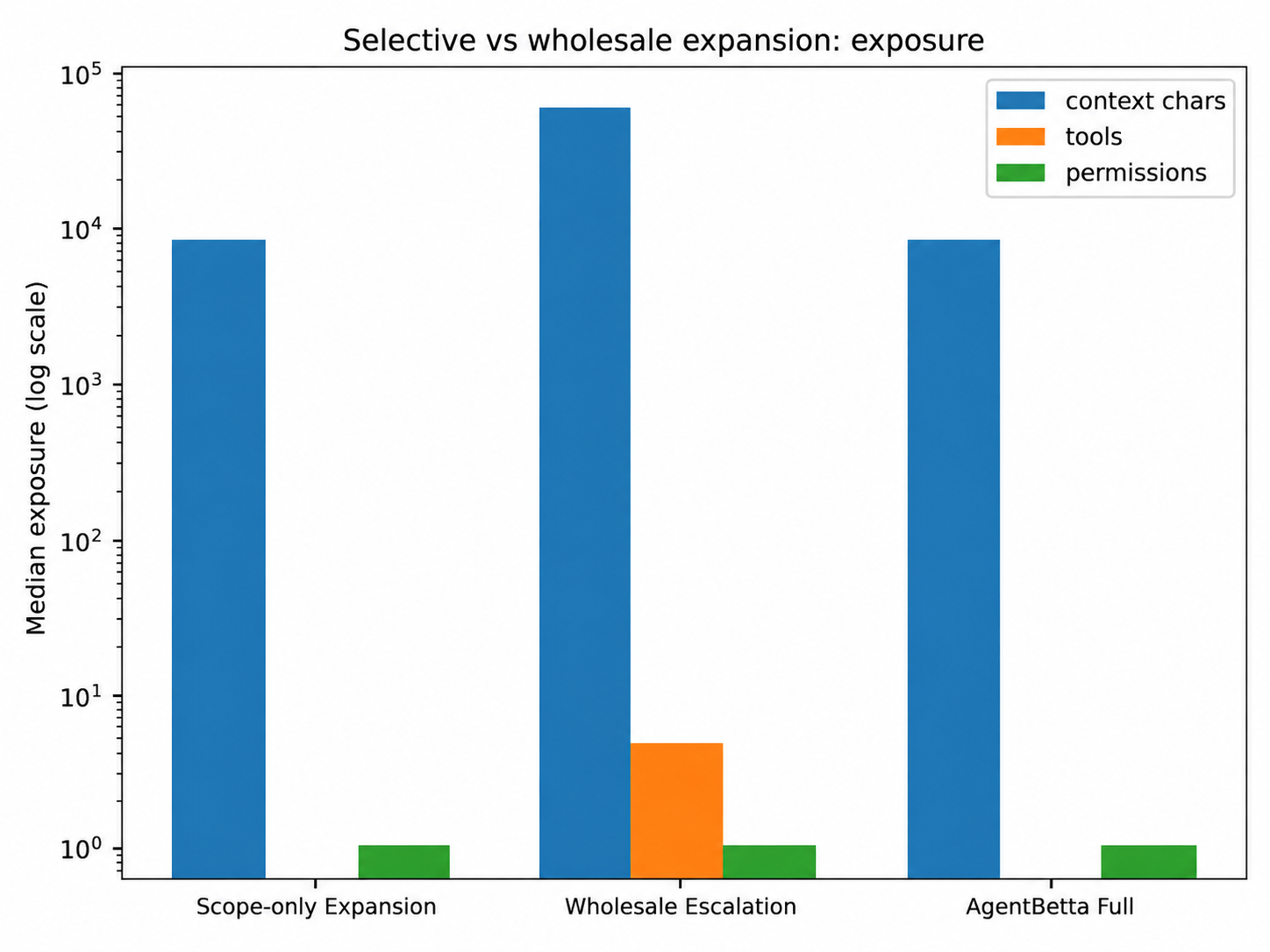}
\caption{Median context, tool, and permission exposure for AgentBetta, scope-only expansion, and wholesale escalation. Wholesale escalation increases dimensions unrelated to the diagnosed deficiency, whereas the selective methods retain the lean exposure profile for dimensions that do not require expansion.}
\label{fig:selective_wholesale}
\end{figure}

The effect of diagnosis on configuration change is summarized in Table~\ref{tab:selective_vs_wholesale}. Among initially failed tasks, AgentBetta recovered 89.0\% with a median of one changed dimension per adaptation and an unrelated-dimension change rate of 0.000. Scope-only expansion produced a similar recovery rate and the same narrow change pattern. Wholesale Escalation recovered 91.67\%, but changed all eight configuration dimensions at the median, produced an unrelated-dimension change rate of 0.733, added 56,000 context units, and exposed five additional tools. Figure~\ref{fig:selective_wholesale} illustrates the corresponding exposure profiles. The controlled advantage of selective expansion is therefore narrower intervention, rather than a higher maximum success rate.

\begin{table}[t]
\centering
\caption{Recovery after initial failure under selective and wholesale adaptation.}
\label{tab:selective_vs_wholesale}
\small
\resizebox{\linewidth}{!}{%
\begin{tabular}{lrrrrrr}
\toprule
Method & Initially failed & Recovery (\%) & Mean adaptations per initially failed task & Median dimensions changed & Unrelated-change rate & Added context \\
\midrule
AgentBetta Full & 482 & 89.00 & 0.814 & 1 & 0.000 & 0 \\
Scope-only Expansion & 192 & 89.06 & 0.947 & 1 & 0.000 & 0 \\
Wholesale Escalation & 192 & 91.67 & 0.909 & 8 & 0.733 & 56,000 \\
\bottomrule
\end{tabular}%
}
\end{table}

\subsection{Verification-gated counterfactual contraction and history reuse}
\label{subsec:contraction_history_results}

\begin{figure}[t]
\centering
\includegraphics[width=\linewidth]{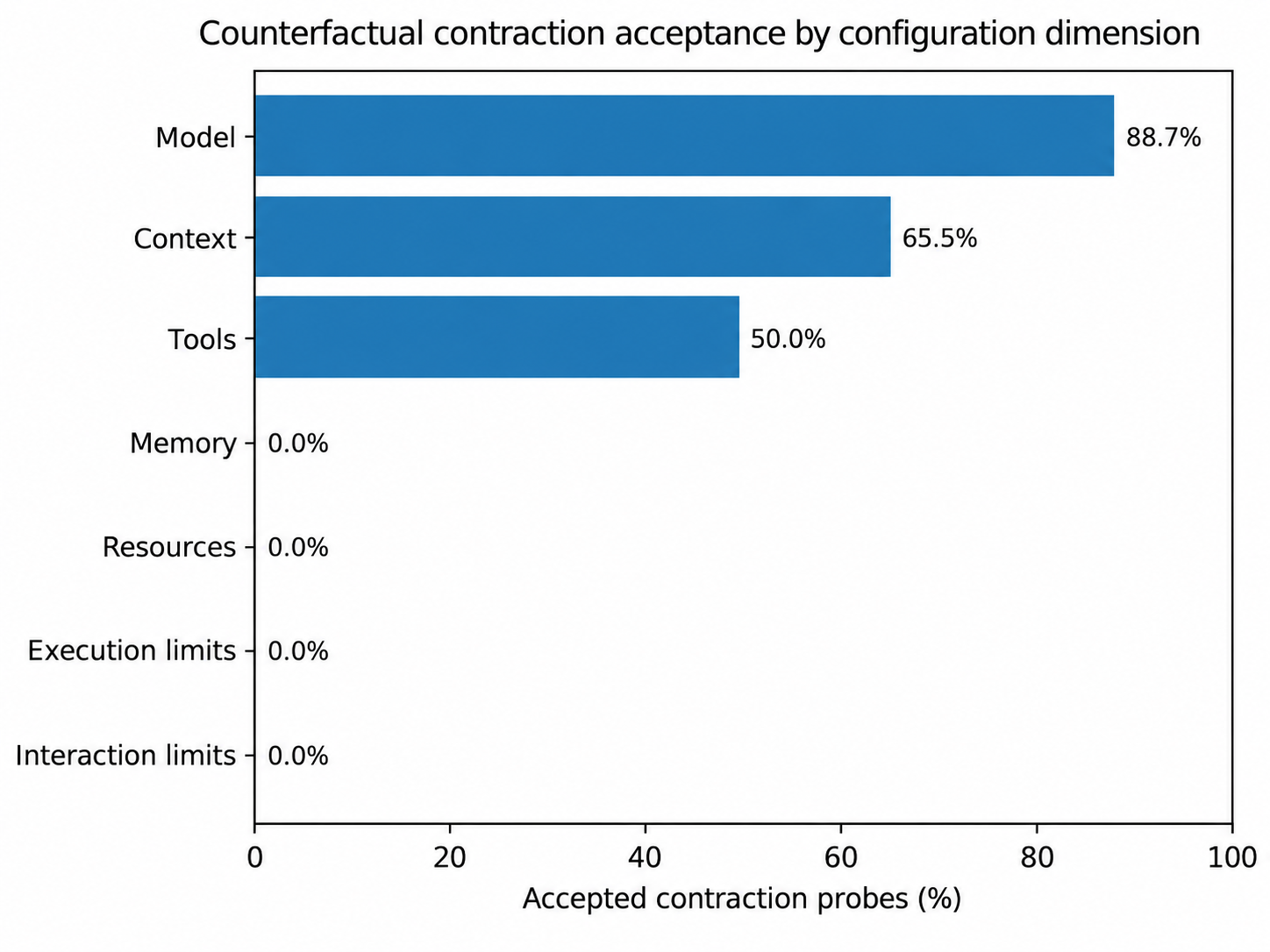}
\caption{Acceptance rate of verification-gated counterfactual contraction by configuration dimension. A contraction is accepted only when re-execution under the reduced configuration remains verified.}
\label{fig:contraction_acceptance}
\end{figure}

Counterfactual contraction was applied only after verified success. Across 894 eligible successful parent runs, 2,393 one-dimension probes were executed; 1,350 retained verification and 1,043 failed, giving an overall acceptance rate of 56.41\%. The outcome was dimension dependent (Table~\ref{tab:contraction}). Model reduction was accepted in 88.70\% of 1,566 probes, context reduction in 65.55\% of 357 probes, and tool reduction in 50.00\% of 156 probes. The evaluated probes did not accept reductions of memory, resource budget, execution limits, or interaction limits. No permission probe was represented in the contraction table; permission behavior was assessed separately by the safety tests. Figure~\ref{fig:contraction_acceptance} presents the acceptance profile by dimension.

\begin{table}[t]
\centering
\caption{Counterfactual contraction outcomes by configuration dimension.}
\label{tab:contraction}
\small
\begin{tabular}{lrrr}
\toprule
Dimension & Probes attempted & Probes passed & Acceptance (\%) \\
\midrule
Model & 1,566 & 1,389 & 88.70 \\
Context & 357 & 234 & 65.55 \\
Tools & 156 & 78 & 50.00 \\
Memory & 15 & 0 & 0.00 \\
Resources & 103 & 0 & 0.00 \\
Execution limits & 103 & 0 & 0.00 \\
Interaction limits & 93 & 0 & 0.00 \\
\midrule
All evaluated probes & 2,393 & 1,350 & 56.41 \\
\bottomrule
\end{tabular}
\end{table}

History-informed initialization produced a narrower effect than diagnosis. On 147 training tasks, both cold and history-informed AgentBetta conditions verified 131 tasks. Median adaptations decreased from 1 to 0 and mean attempts decreased from 1.88 to 1.56, whereas median starting context remained 8,000 and median starting tool exposure remained zero. In the present implementation, history therefore reduced repeated search but did not reduce initial exposure or improve task success. This negative result is consistent with the stored evidence: the initializer retrieves previously verified configurations, which establish sufficiency but do not by themselves establish that a smaller starting configuration would also have succeeded.

\subsection{External comparative results}
\label{subsec:external_summary}

Table~\ref{tab:external_summary} summarizes the task-level comparisons. The external results do not support a single ordering across task classes. AgentBetta exceeded matched ReAct on BFCL, but required more inference. It achieved higher verified success than the evaluated AutoGen and CrewAI configurations on the complete BFCL framework subset while using fewer tokens and model calls. The GAIA difference relative to Reflexion was unresolved and the repeated panel remained incomplete. The SWE-bench pilot showed a substantially higher observed task-resolution rate for SWE-agent, although the Holm-adjusted test did not meet the conventional 0.05 threshold. On an additional local model family, all four tested controllers reached the same accuracy on the small cross-family subset.

\begin{table*}[t]
\centering
\caption{Summary of external comparisons. Resource effects are AgentBetta relative to the comparator and are computed from task-level repetition aggregates. ``Adj. $p$'' denotes the Holm-adjusted significance level within the predeclared confirmatory family.}
\label{tab:external_summary}
\scriptsize
\resizebox{\textwidth}{!}{%
\begin{tabular}{llrrrrll}
\toprule
Comparator & Benchmark & AgentBetta (\%) & Comparator (\%) & Difference (pp) & 95\% CI (pp) & Adj. $p$ & Principal resource effect \\
\midrule
ReAct & BFCL & 87.35 & 83.94 & +3.40 & $[+0.09,+6.89]$ & 0.0103 & tokens +16.9\%; calls +20.2\%; cost +18.1\% \\
Reflexion & GAIA & 27.19 & 31.58 & $-4.39$ & $[-15.35,+5.70]$ & 0.4531 & tokens $-41.0$\%; calls $-31.1$\%; cost $-48.9$\% \\
AutoGen & BFCL subset & 88.02 & 31.77 & +56.25 & $[+45.3,+66.7]$ & $5.8\times10^{-11}$ & tokens $-66.4$\%; calls $-53.0$\%; cost $-70.5$\% \\
CrewAI & BFCL subset & 88.02 & 13.54 & +74.48 & $[+65.1,+83.3]$ & $8.9\times10^{-15}$ & tokens $-77.0$\%; calls $-67.7$\%; cost $-81.2$\% \\
SWE-agent & SWE-bench pilot & 6.25 & 75.00 & $-68.75$ & $[-93.8,-37.5]$ & 0.0625 & tokens $-79.4$\%; calls $-35.2$\%; cost $-76.4$\% \\
Fixed Full & Cross-family BFCL & 31.25 & 31.25 & 0.00 & $[0.0,0.0]$ & -- & context $-75.0$\%; tokens +134.9\% \\
\bottomrule
\end{tabular}%
}
\end{table*}

\subsubsection{Matched ReAct comparison on BFCL}
\label{subsubsec:bfcl_react}

\begin{figure}[t]
\centering
\includegraphics[width=\linewidth]{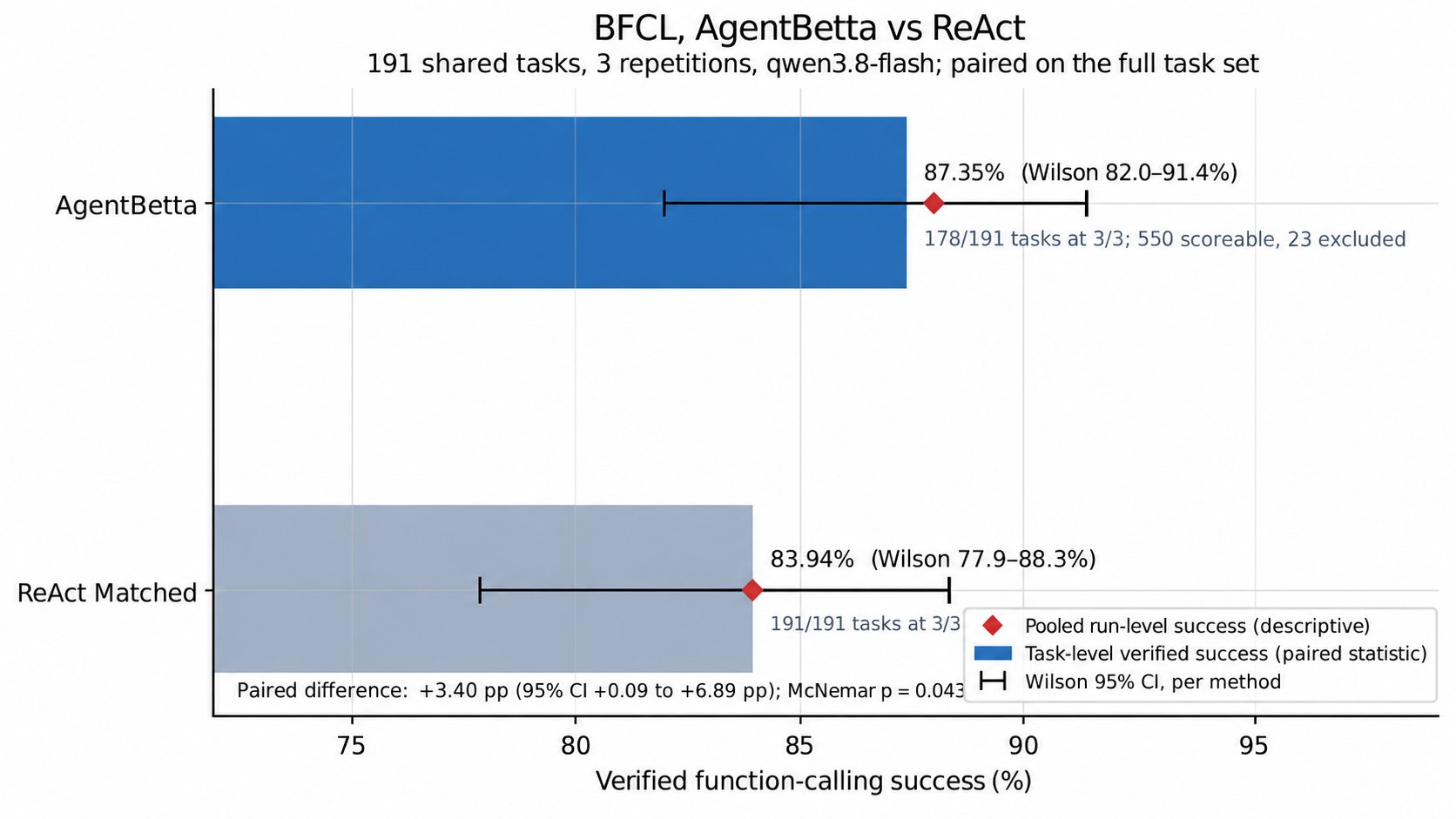}
\caption{Task-level verified success for AgentBetta and matched ReAct on the repeated BFCL panel. AgentBetta's incomplete repetition coverage is reported explicitly rather than imputed.}
\label{fig:bfcl_success_ci}
\end{figure}

BFCL evaluates an LLM's ability to select and emit valid function calls under a range of tool-use conditions \cite{patil2025bfcl}. The matched ReAct comparison \cite{yao2023react} used 191 shared tasks with three intended repetitions. ReAct completed all 191 tasks with three scoreable repetitions. AgentBetta had complete three-repetition coverage on 178 tasks; 23 repetitions were excluded because provider failures could not be recovered under the frozen provider condition. Table~\ref{tab:bfcl_react} reports the repeated-panel coverage and resource summary. AgentBetta achieved 87.35\% task-level verified success and ReAct 83.94\%, giving a difference of $+3.40$ percentage points (95\% CI $[+0.09,+6.89]$; raw $p=0.0034$; Holm-adjusted $p=0.0103$). Figure~\ref{fig:bfcl_success_ci} reports the success estimates and coverage-aware intervals.

\begin{figure}[t]
\centering
\includegraphics[width=\linewidth]{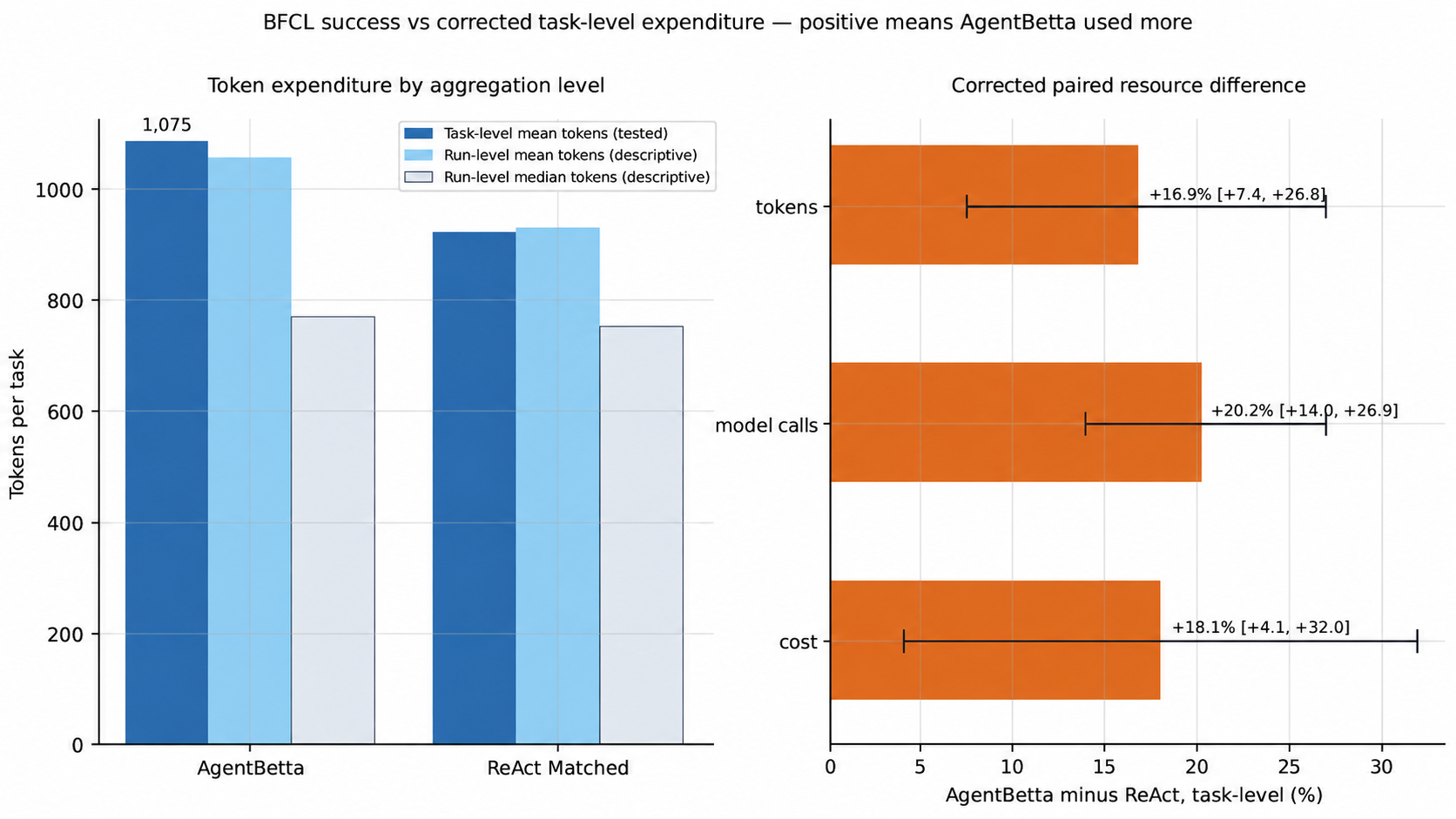}
\caption{BFCL success and task-level inference expenditure for AgentBetta and matched ReAct. AgentBetta attains higher verified success in this panel but uses more tokens and model calls.}
\label{fig:bfcl_expenditure}
\end{figure}

The success improvement was accompanied by greater inference expenditure. After the repeated-measures correction, AgentBetta used 16.85\% more tokens, 20.24\% more model calls, and 18.08\% higher model cost than ReAct. Task-level context and tool exposure did not differ reliably. Figure~\ref{fig:bfcl_expenditure} therefore represents the BFCL result as a success--expenditure trade-off rather than as a joint accuracy-and-efficiency improvement.

\begin{table*}[t]
\centering
\caption{Repeated BFCL comparison with matched ReAct. Success and resource values are task-level repetition aggregates; pooled run-level success is included only as a descriptive quantity.}
\label{tab:bfcl_react}
\scriptsize
\resizebox{\textwidth}{!}{%
\begin{tabular}{lrrrrrrrr}
\toprule
Method & Scoreable runs & Excluded runs & Tasks complete 3/3 & Task-level success (\%) & Pooled success (\%) & Mean tokens/task & Mean calls/task & Mean cost/task (USD) \\
\midrule
AgentBetta & 550 & 23 & 178/191 & 87.35 & 88.00 & 1,075 & 1.20 & 0.000262 \\
ReAct Matched & 573 & 0 & 191/191 & 83.94 & 83.94 & 920 & 1.00 & 0.000222 \\
\bottomrule
\end{tabular}%
}
\end{table*}

\subsubsection{Evaluated AutoGen and CrewAI configurations}
\label{subsubsec:framework_comparison}

The multi-agent comparison used a frozen 64-task BFCL subset with three complete repetitions for each method. AutoGen provides conversational multi-agent orchestration \cite{wu2024autogen}, whereas CrewAI provides role-based agent and workflow abstractions \cite{crewai2026}. Under the evaluated configurations, AgentBetta reached 88.02\% verified success, compared with 31.77\% for AutoGen and 13.54\% for CrewAI. The absolute differences were $+56.25$ and $+74.48$ percentage points, respectively, and both remained significant after Holm correction (Table~\ref{tab:external_summary}).

\begin{figure}[t]
\centering
\includegraphics[width=\linewidth]{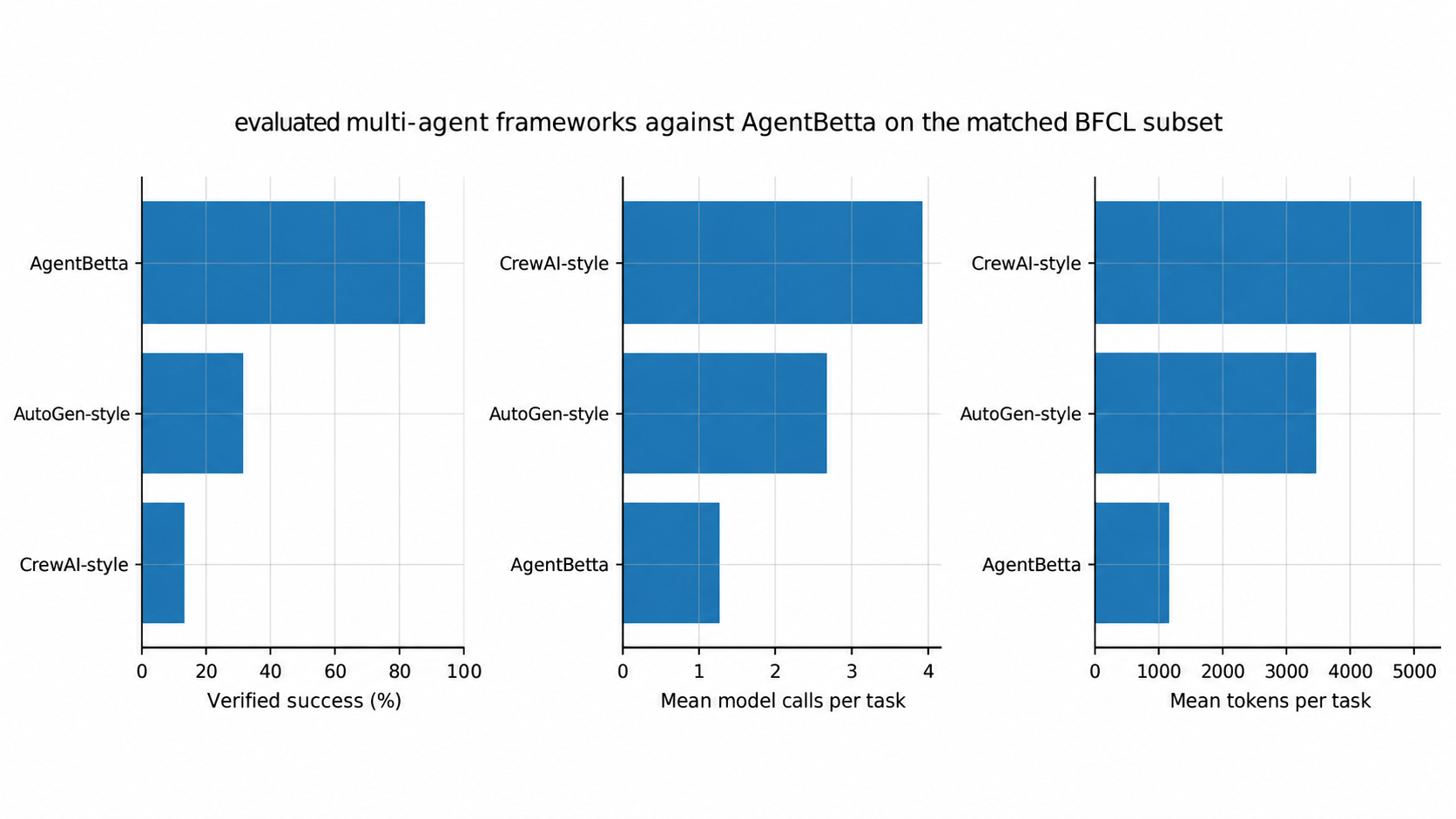}
\caption{Verified success and inference overhead for AgentBetta and the evaluated AutoGen and CrewAI configurations on the complete 64-task BFCL framework subset.}
\label{fig:framework_overhead}
\end{figure}

The comparison also favored AgentBetta on inference expenditure. Relative to AutoGen, AgentBetta used 66.38\% fewer tokens, 52.99\% fewer model calls, and 70.47\% lower model cost. Relative to CrewAI, the corresponding reductions were 76.95\%, 67.68\%, and 81.19\%. Figure~\ref{fig:framework_overhead} summarizes the success and inference profiles. These values characterize the tested framework configurations on the common BFCL subset; they do not constitute a framework-wide ranking independent of orchestration design, task decomposition, or model choice.

\subsubsection{Reflexion comparison on GAIA}
\label{subsubsec:gaia_reflexion}

GAIA evaluates general-assistant tasks requiring combinations of reasoning, information access, and tool use \cite{mialon2024gaia}. The text-only comparison with Reflexion \cite{shinn2023reflexion} used a 40-task panel with three intended repetitions. AgentBetta had 115 scoreable runs and complete 3/3 coverage on 37 tasks; Reflexion had 65 scoreable runs and complete coverage on 11 tasks. Fifty-five Reflexion repetitions and five AgentBetta repetitions were unscorable. The unrecovered Reflexion set included provider failures and runs in which the shared model-call allowance was exhausted before the evaluation judge completed scoring. The missing repetitions could not be rerun under the identical frozen provider configuration; they were therefore retained as unscored rather than reproduced under a different provider condition.

\begin{figure}[t]
\centering
\includegraphics[width=\linewidth]{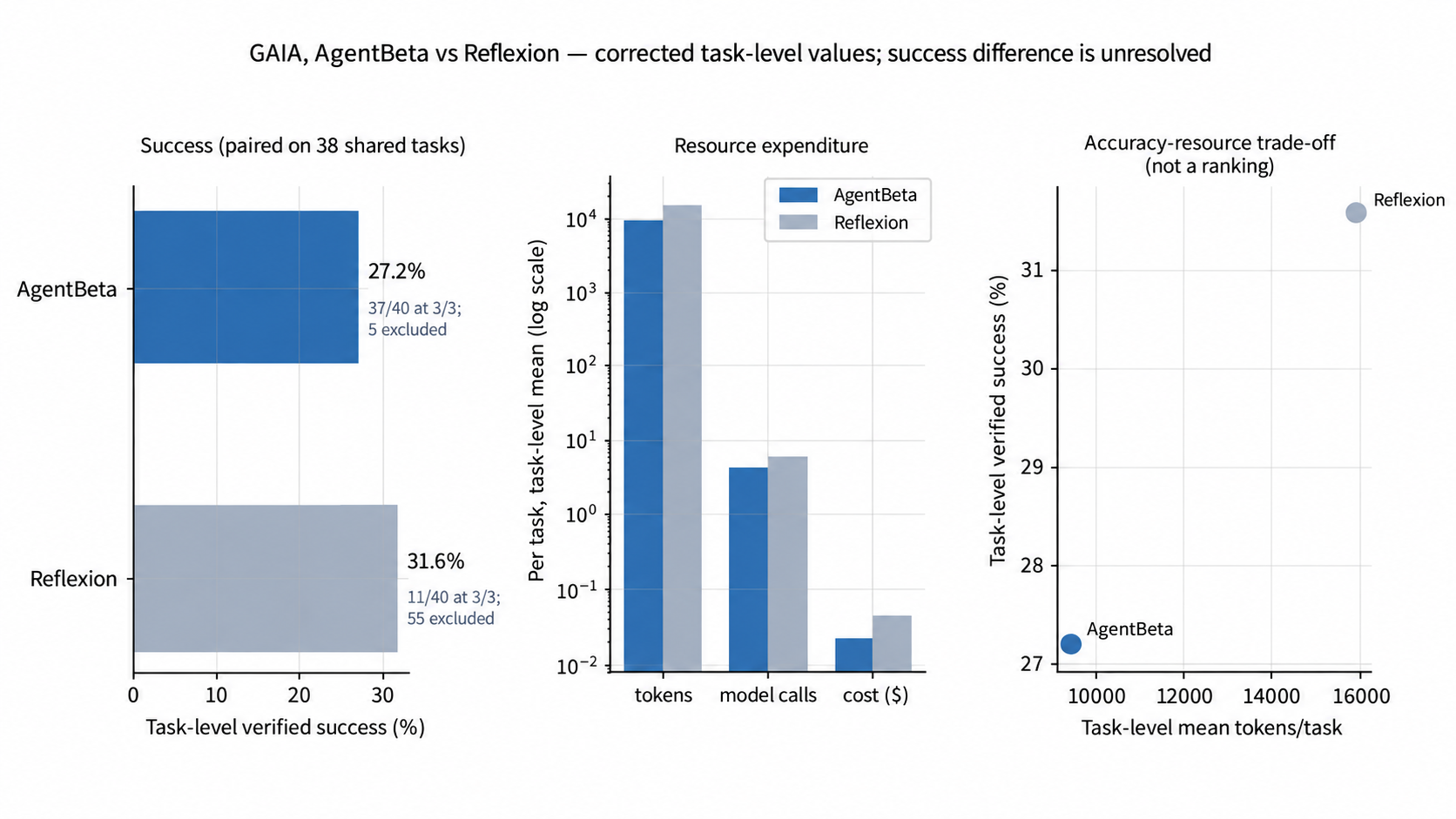}
\caption{Task-level GAIA success and resource expenditure for AgentBetta and Reflexion. The figure is descriptive because the repeated panel is incomplete, particularly for Reflexion.}
\label{fig:gaia_tradeoff}
\end{figure}

On the tasks with scoreable data for both methods, AgentBetta achieved 27.19\% task-level verified success and Reflexion 31.58\%, a difference of $-4.39$ percentage points (95\% CI $[-15.35,+5.70]$; raw and Holm-adjusted $p=0.4531$). The interval crosses zero, so the success difference is unresolved. AgentBetta used 41.01\% fewer tokens, 31.06\% fewer model calls, and 48.92\% lower model cost. Context exposure was 5.26\% higher and not statistically reliable; tool exposure did not differ. Table~\ref{tab:gaia_reflexion} and Fig.~\ref{fig:gaia_tradeoff} therefore support an exploratory accuracy--resource trade-off, not a performance ranking.

\begin{table*}[t]
\centering
\caption{GAIA comparison between AgentBetta and Reflexion. The repeated panel remains incomplete because frozen-provider recovery was unavailable; the result is therefore interpreted as exploratory.}
\label{tab:gaia_reflexion}
\scriptsize
\resizebox{\textwidth}{!}{%
\begin{tabular}{lrrrrrrrr}
\toprule
Method & Scoreable runs & Excluded runs & Tasks complete 3/3 & Task-level success (\%) & Pooled success (\%) & Mean tokens/task & Mean calls/task & Mean cost/task (USD) \\
\midrule
AgentBetta & 115 & 5 & 37/40 & 27.19 & 26.96 & 9,364 & 4.47 & 0.003189 \\
Reflexion & 65 & 55 & 11/40 & 31.58 & 38.46 & 15,874 & 6.48 & 0.006243 \\
\bottomrule
\end{tabular}%
}
\end{table*}

\subsubsection{Specialized coding-agent comparison}
\label{subsubsec:swe_results}

The coding pilot used eight screened SWE-bench Verified instances \cite{jimenez2024swebench}, with two repetitions and a hard model-call ceiling. SWE-agent, whose agent--computer interface is designed specifically for repository navigation, editing, and test execution \cite{yang2024sweagent}, solved 75.00\% of the evaluated runs. AgentBetta solved 6.25\% and matched ReAct solved 0\%. The AgentBetta--SWE-agent difference was $-68.75$ percentage points with a 95\% interval of approximately $[-93.8,-37.5]$ percentage points. The raw paired significance level was 0.0312, but the Holm-adjusted value was 0.0625. The observed difference is therefore large, but the small pilot does not establish confirmatory significance after multiplicity correction.

\begin{figure}[t]
\centering
\includegraphics[width=\linewidth]{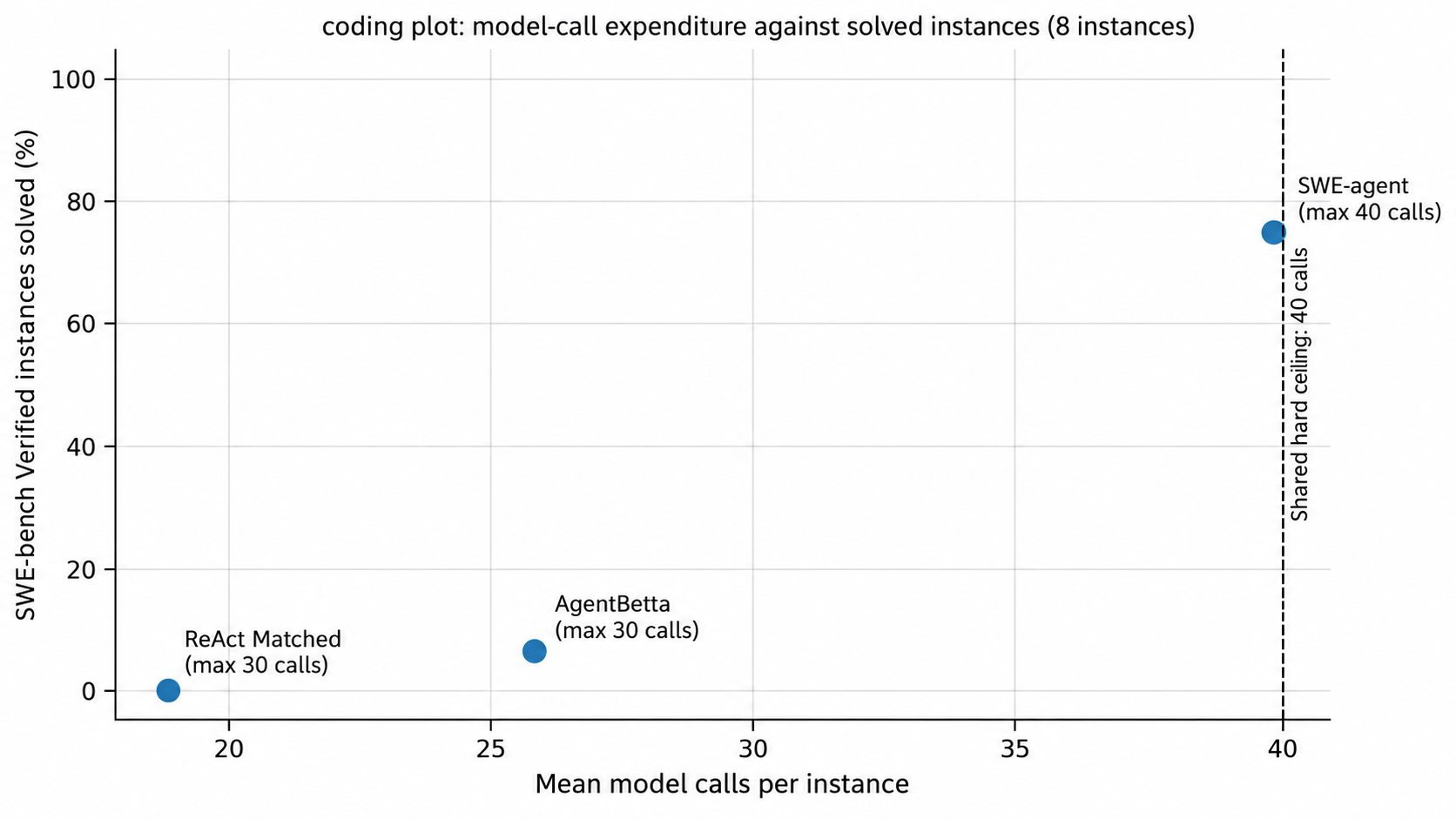}
\caption{Observed success and execution consumption in the eight-instance SWE-bench pilot. The specialized SWE-agent configuration achieved substantially higher task resolution, while AgentBetta consumed fewer inference resources.}
\label{fig:swe_pilot}
\end{figure}

AgentBetta used 79.38\% fewer tokens, 35.16\% fewer model calls, 76.41\% lower model cost, and 88.00\% less configured context than SWE-agent, yet this lower expenditure coincided with substantially lower task resolution. Figure~\ref{fig:swe_pilot} makes this asymmetry explicit. The pilot supports the continued value of domain-specific execution scaffolding for repository-scale software engineering.

\subsubsection{Cross-family replication}
\label{subsubsec:cross_family}

\begin{figure}[t]
\centering
\includegraphics[width=\linewidth]{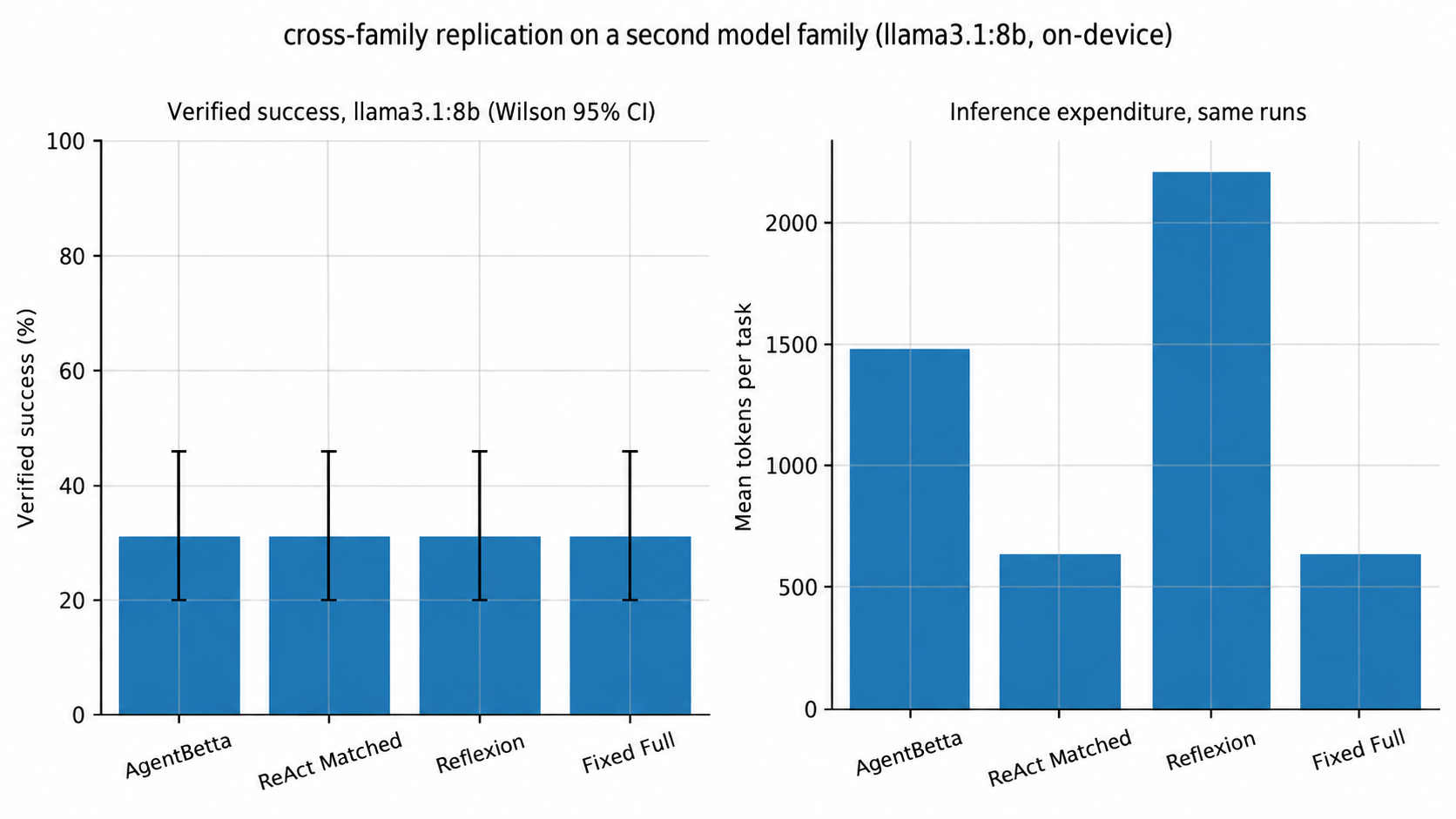}
\caption{Cross-family BFCL replication on the 16-task stratified subset. All four controllers reached the same verified success, so the main-backbone accuracy ordering did not replicate.}
\label{fig:cross_family}
\end{figure}

A 16-task stratified BFCL subset was repeated on an additional local model family. AgentBetta, Fixed Full, matched ReAct, and Reflexion each achieved 31.25\% verified success. The accuracy ordering observed with the primary backbone therefore did not reproduce on this additional family. AgentBetta used less configured context than Fixed Full but more tokens and model calls than ReAct; it used fewer tokens and calls than Reflexion. Figure~\ref{fig:cross_family} summarizes these results. The experiment supports portability of the runtime and configuration protocol, but it does not support a claim of model-independent performance superiority.

\subsection{Safety and permission behavior}
\label{subsec:safety_results}

The controlled campaign recorded zero unauthorized permission activations and zero successful hard-denied operations. The verifier also rejected adversarial model text when structured execution evidence indicated failure, and history-based initialization was unable to widen the current task's authority envelope. These checks establish that capability prediction and authorization were separated in the evaluated implementation.

The safety result must nevertheless be interpreted at the correct level. In the external comparison, all methods executed through a shared enforcement boundary. Consequently, zero successful hard-denied dispatches are evidence for the common permission layer rather than for AgentBetta alone. AgentBetta-specific evidence concerns the explicit representation of permission as a configuration dimension, the requirement that expansion remain inside the eligible authority set, and the ability to test whether capability can be reduced after verification.

\subsection{Results synthesis}
\label{subsec:results_summary}

The controlled experiments support three mechanism-level conclusions. First, diagnosis is the principal component responsible for recovering from insufficient initial configurations. Second, selective adaptation can preserve most of the recovery obtained by wholesale escalation while avoiding unrelated changes to the configuration vector. Third, counterfactual contraction can remove capability after success in a substantial fraction of the evaluated probes, although its effect is on verified exposure rather than on task success. History reuse reduces repeated adaptation work but does not yet compress the starting configuration.

The external comparisons provide a more conditional result. AgentBetta achieved a modest, statistically supported BFCL success increase over matched ReAct but used more inference. It achieved higher verified success than the evaluated AutoGen and CrewAI configurations on a complete BFCL subset while consuming fewer tokens, model calls, and cost. The GAIA comparison with Reflexion remained unresolved and incomplete. The SWE-bench pilot favored the specialized SWE-agent in observed task resolution, and the cross-family panel did not reproduce the primary-backbone accuracy ordering. Collectively, these results support AgentBetta as a task-conditioned capability-configuration controller rather than as a universally dominant agent architecture.

\section{Discussion}
\label{sec:discussion}

\subsection{Configuration adaptation as a runtime control problem}
\label{subsec:discussion_control}

The controlled results indicate that treating configuration as a runtime variable can improve capability allocation under the evaluated conditions. A lean fixed configuration fails when the task requires capability outside its initial envelope, whereas a full fixed configuration avoids those failures by provisioning the maximum tested capability irrespective of task need. AgentBetta instead uses an adaptive configuration regime that begins from a bounded configuration, diagnoses insufficiency from execution and verification evidence, and modifies only the relevant dimensions. Accordingly, the contribution is not the assumption that smaller configurations are inherently preferable, but the mechanism for adapting capability according to task evidence as execution proceeds.

This interpretation should be distinguished from broader claims of adaptive agent configuration. Recent systems have explored query-conditioned configuration, complexity-aware execution, and dynamic sub-agent orchestration \cite{taparia2026arc,yin2026e3,ruan2026aorchestra}. The present results do not support novelty for adaptation in the broad sense. They support the narrower mechanism studied here: diagnosis of deficient coordinates in a heterogeneous configuration vector, selective expansion constrained by an authority envelope, verification of the resulting execution, and post-success testing of whether individual coordinates can be reduced.

\subsection{Why diagnosis dominates the observed success effect}
\label{subsec:discussion_diagnosis}

The ablation results identify diagnosis as the component most directly associated with successful recovery. The macro-F1 score of 0.819 is particularly informative when considered together with a precision of 1.000 across all evaluated dimensions: the controller rarely broadens the configuration without evidence for the corresponding deficiency. When diagnosis is removed, the adaptive loop loses the mechanism that maps failure evidence to a specific intervention and verified success falls to the level of non-adaptive query-conditioned execution. The selective-expansion experiment provides a second view of the same mechanism: similar recovery is obtained with one changed dimension rather than eight.

The lower tool recall also defines an empirical boundary. A controller cannot reliably diagnose a requirement that is absent from both the task representation and the available execution evidence. This limitation does not justify indiscriminate pre-emptive expansion; rather, it indicates that diagnosis accuracy depends on observable evidence. In deployment, richer tool descriptions, structured failure signals, or explicit uncertainty about hidden requirements are likely to be more productive than indiscriminate escalation.

\subsection{Counterfactual contraction serves a different objective from recovery}
\label{subsec:discussion_contraction}

Contraction did not materially increase verified task success, and it should not be presented as a success mechanism. Its role begins after successful execution. Overall, 56.41\% of the evaluated single-dimension contraction probes preserved verification, indicating that removable capability was present in some successful configurations under the tested conditions. Model, context, and tool reductions were accepted at different rates, indicating that redundant capacity is not distributed uniformly across configuration dimensions.

This function is related to recent work on reducing or adapting agent workflows after execution, including AgentSlimming and counterfactual harness adaptation \cite{chen2026agentslimming,fu2026chill}. The distinction relevant to AgentBetta is the object being tested: each probe alters a coordinate of the heterogeneous execution configuration and is accepted only after re-execution remains verified. The present method does not establish a globally minimal configuration. Because probes are local and largely dimension-wise, interaction effects between coordinates may remain undiscovered. The appropriate claim is verified empirical reduction, not global optimality.

\subsection{History currently reduces search rather than initial exposure}
\label{subsec:discussion_history}

The history experiment yielded no improvement in verified success or initial capability exposure, although it reduced repeated adaptation effort. Reusing prior verified configurations reduced median adaptations from 1 to 0 and mean attempts from 1.88 to 1.56. This outcome is explained by the evidence available to the initializer. A previously successful configuration proves sufficiency; unless the historical record also contains evidence that smaller alternatives were verified, it provides no basis for starting smaller.

A more informative history mechanism would therefore combine successful configurations with the outcomes of contraction probes. Accepted reductions identify configurations known to remain sufficient after capability removal, while failed reductions identify lower bounds for specific task families. Learning from both directions could turn history from a search shortcut into a source of empirically supported initialization policies. This remains a testable extension rather than a result of the current implementation.

\subsection{External comparisons reveal task-dependent trade-offs}
\label{subsec:discussion_external}

The BFCL comparison demonstrates that configuration adaptation can improve verified function-calling performance without necessarily reducing inference. AgentBetta's 3.40 percentage-point advantage over matched ReAct was statistically supported, but it required more tokens and model calls. ReAct modifies the reasoning--action trajectory \cite{yao2023react}; AgentBetta modifies the capability envelope within which that trajectory is executed. The evaluated BFCL result is therefore consistent with complementary intervention points rather than with a general replacement relationship.

The GAIA comparison leads to a different conclusion. Reflexion explicitly uses verbal reinforcement and reflective retry \cite{shinn2023reflexion}; AgentBetta used substantially fewer resources on the paired scoreable tasks, but the success difference was unresolved and the repetition coverage was incomplete. For tasks dominated by reasoning revision rather than by missing capability, reflection may be more consequential than configuration change. The current GAIA evidence is insufficient to rank the two approaches and is retained as an exploratory trade-off until the frozen missing runs can be recovered.

The multi-agent comparison provides clearer evidence within the evaluated experimental scope. On the complete BFCL subset, the evaluated AutoGen and CrewAI configurations consumed more inference while obtaining lower verified success. Multi-agent systems are designed to support coordination, role specialization, and decomposition \cite{wu2024autogen,crewai2026}; BFCL does not require all of those properties. The result therefore supports a limited conclusion: when a task can be solved within a single execution path, a dynamically configured single agent can avoid coordination overhead that is unnecessary for that task. It does not establish that multi-agent collaboration is inferior on problems that genuinely benefit from distributed roles or parallel decomposition.

\subsection{Specialized procedural scaffolding remains necessary}
\label{subsec:discussion_specialization}

The SWE-bench pilot shows that adaptive configuration does not substitute for domain-specific execution scaffolding. SWE-agent was designed around an agent--computer interface specialized for software engineering \cite{yang2024sweagent}; its substantially higher observed task resolution indicates that selecting resources and permissions is not equivalent to supplying domain-specific procedures for codebase navigation, editing, test execution, and patch refinement. AgentBetta consumed fewer resources, but the resulting economy did not compensate for the task-performance gap.

This observation suggests a practical extension of the configuration view. A specialized executor can itself be treated as a selectable capability. AgentBetta need not reproduce every procedural strategy internally; instead, the configuration controller could route a task to a coding-specific, research-specific, or other specialized executor when the diagnosed task structure requires it. Such a design would preserve task-conditioned capability allocation while acknowledging that configuration control and domain expertise address different sources of failure.

\subsection{Model dependence limits general performance claims}
\label{subsec:discussion_generalization}

The cross-family experiment did not reproduce the primary-backbone ranking: all tested controllers reached 31.25\% on the 16-task subset. This result constrains any claim that an adaptive controller will improve accuracy independently of the underlying model. Configuration determines what capability is made available; it does not create reasoning or function-calling competence that the backbone lacks. A controller can therefore reduce avoidable exposure or allocate resources more appropriately while still failing to change the dominant error source.

The current evidence supports portability of the runtime mechanism across model families, not invariant performance gain. A stronger generalization claim would require larger matched panels across independent providers and model families, with the same task identifiers, tool definitions, verifier, and resource ceilings. Until such evidence is available, performance claims should remain conditional on the tested model and benchmark configuration.

\subsection{Safety implications of bounded capability expansion}
\label{subsec:discussion_safety}

Treating permission as an explicit configuration dimension creates a useful separation between predicted need and granted authority. The controller may infer that an operation would help complete a task, but the requested expansion remains bounded by the externally defined eligible set. This property is consistent with least-authority execution and prevents history-based initialization from silently granting broader rights than the current policy permits.

The evaluation does not establish a complete security guarantee. The shared enforcement layer prevented prohibited dispatches for all compared agents, so those outcomes cannot be attributed uniquely to AgentBetta. The agent-specific contribution is the explicit inclusion of authority in configuration selection and adaptation, together with the ability to retain or reduce authority based on task evidence. Adversarial evaluation with more complex tool chains, indirect prompt injection, and persistent cross-task state remains necessary before stronger security claims can be made.

\subsection{Threats to validity}
\label{subsec:discussion_validity}

Several limitations bound the interpretation of the results. First, AB-ConfigBench couples task requirements to a deterministic configuration-aware provider. This is appropriate for isolating controller behavior, but absolute success in that layer should not be interpreted as general language-model performance. The external evaluation was included specifically to test whether the mechanism remains meaningful under model inference.

Second, the repeated external campaign is incomplete for BFCL AgentBetta and especially for GAIA Reflexion. Twenty-three of 573 intended BFCL AgentBetta attempts and 60 of 240 intended GAIA attempts across AgentBetta and Reflexion remained unscored because the frozen provider could not be used for recovery. AgentBetta retained full 3/3 BFCL coverage on 178 of 191 tasks; Reflexion retained full 3/3 GAIA coverage on only 11 of 40 tasks. The BFCL effect is therefore better supported than the GAIA comparison. The GAIA result is reported as exploratory and should be updated if the frozen missing runs are recovered.

Third, eight excluded GAIA Reflexion runs exhausted the same model-call allowance subsequently needed by the evaluation judge. Future campaigns should isolate the benchmark judge's allowance from the agent's execution budget so that an agent's internal call consumption cannot determine whether its final output is scoreable. Fourth, the SWE-bench experiment contains only eight screened instances and uses a local reproduction rather than a large official-environment campaign. Although the observed effect is large, the Holm-adjusted significance level exceeds 0.05.

Fifth, wall-clock time is confounded by provider retries and network behavior and is therefore excluded from the principal efficiency claims. Tokens, model calls, model cost, and configuration exposure provide more reproducible comparisons in the present data. Sixth, the cross-family replication contains only 16 BFCL tasks and one additional local model family. Finally, contraction is local and dimension-wise; neither the algorithm nor the empirical results establish a globally minimal configuration.

\subsection{Implications for adaptive agent design}
\label{subsec:discussion_implications}

The combined evidence suggests that efficiency in agent systems should not be represented by a single scalar objective. AgentBetta used more inference than ReAct on BFCL, less inference than the evaluated multi-agent configurations, and far less inference than SWE-agent while also performing substantially worse on the coding pilot. The results therefore motivate evaluating whether the resources, tools, authority, and procedural mechanisms allocated to a task are justified by verified performance.

Within this interpretation, AgentBetta contributes an explicit control layer for heterogeneous capability. The controlled experiments show that high-precision diagnosis can support selective expansion, and that verified configurations can subsequently be tested for reducible capacity. The external experiments show that the value of this control is task dependent: it can improve function-calling performance, compare favorably with the evaluated multi-agent configurations on tasks solvable through a single execution path, remain unresolved against reflective reasoning, and be insufficient when specialized procedural scaffolding dominates. Accordingly, the results indicate that adaptive capability configuration complements reasoning, reflection, collaboration, and specialization rather than replacing them.

\section{Conclusion}
\label{sec:conclusion}

This study developed and evaluated AgentBetta, a verification-driven adaptive AI Nano-Agent that represents an agent's executable state as a joint configuration of model capability, context, tools, permissions, memory, computational resources, execution bounds, and interaction limits. The central objective was not to minimize a single resource in isolation, but to determine whether heterogeneous capability can be allocated, expanded, and reduced in response to task evidence while preserving verifiable task completion. To this end, AgentBetta combines task-conditioned initialization, configuration-deficiency diagnosis, dimension-selective expansion, runtime verification, controlled counterfactual contraction, and empirical reuse of prior verified configurations within a single auditable control loop.

The controlled evaluation showed that the main performance contribution arises from diagnosis-guided adaptation rather than from broad escalation. AgentBetta achieved 91.38\% verified success, compared with 93.10\% for Fixed Full, while using substantially lower median context and tool exposure. Diagnosis achieved high precision across all configuration dimensions, and selective expansion avoided unrelated configuration changes that were common under wholesale escalation. Counterfactual contraction further demonstrated that some successful configurations contained removable capability under the evaluated conditions: 56.41\% of the evaluated one-dimension reductions remained verifiable after re-execution. These results support the use of verification evidence both to identify insufficient capability after failure and to test reducible capability after success. At the same time, the ablation results showed that contraction and history reuse should not be interpreted as independent sources of higher task success under the evaluated conditions. Their primary value was in exposure reduction and adaptation efficiency.

The external comparisons further clarify the role of adaptive configuration. On BFCL, AgentBetta achieved a modest but statistically supported improvement in verified success over matched ReAct, although this gain required greater token, model-call, and cost expenditure. On the evaluated 64-task BFCL subset, AgentBetta achieved higher verified success and lower inference expenditure than the tested AutoGen and CrewAI configurations; this finding is restricted to the evaluated configurations and task setting. By contrast, the GAIA comparison with Reflexion remained statistically unresolved, despite lower inference expenditure for AgentBetta on the paired scoreable tasks. The SWE-bench pilot provided a direct limitation on a universal performance claim: the specialized SWE-agent baseline achieved a substantially higher observed task-resolution rate than AgentBetta on the evaluated repository-scale coding tasks. Similarly, the cross-family experiment did not reproduce the primary-backbone accuracy ordering, indicating that the benefit of adaptive configuration remains dependent on the capability of the underlying model and the evaluated task distribution.

Overall, the findings support interpreting AgentBetta as a configuration-control mechanism whose effects depend on the task, model, and execution environment. Adaptive configuration complements reasoning, reflection, collaboration, and domain-specific procedural scaffolding rather than replacing them. The principal contribution is therefore the treatment of heterogeneous agent capability as an explicit, observable, and revisable control object whose expansion and contraction are conditioned on verification evidence and policy constraints. This formulation provides a practical basis for reducing unjustified capability exposure while retaining the ability to escalate when a task demands additional resources or authority.

Several limitations remain. The controlled benchmark isolates configuration behavior through a deterministic provider and should not be interpreted as a direct measure of open-ended language-model intelligence. The repeated external campaign contains incomplete BFCL and GAIA coverage because some frozen-provider runs could not be recovered, with the GAIA comparison particularly affected. The SWE-bench study used only eight screened instances, and the cross-family replication used a small additional-model panel. Wall-clock latency was also excluded from the principal efficiency claims because provider retry behavior confounded timing measurements. Future work should therefore evaluate AgentBetta across larger independent model families, complete repeated benchmark panels, separate evaluator resources from agent execution budgets, and investigate whether accepted contraction evidence can be used to learn smaller and more reliable initial configurations. A further direction is to treat specialized executors themselves as selectable capabilities, allowing the adaptive controller to route tasks to domain-specific procedures when configuration adjustment alone is insufficient.

% \section*{Data and Code Availability}
% The AgentBetta source code, benchmark definitions, frozen task manifests, evaluation records, and analysis artifacts supporting the reported results are retained in versioned research archives and are available from the corresponding author upon reasonable request.

\section*{Code Availability}
The source code for AgentBetta is publicly available at
\url{https://github.com/ashraful388/AgentBetta}. The repository contains the maintained implementation of the framework \cite{babu2026agentbetta}.
% Before submission, add the exact frozen experimental release tag and commit hash here if these identifiers are available.

% Add funding and competing-interest declarations here according to the
% factual circumstances and the target journal's required format.

\bibliographystyle{unsrtnat}
\bibliography{references} 
\end{document}